\RequirePackage{fix-cm}
\documentclass{article}

\PassOptionsToPackage{numbers, compress}{natbib}
\usepackage[main, final]{neurips_2026}
\usepackage[utf8]{inputenc} 
\usepackage[T1]{fontenc}    
\usepackage{hyperref}       
\usepackage{url}            
\usepackage{booktabs}       
\usepackage{amsfonts}       
\usepackage{nicefrac}       
\usepackage{microtype}      
\usepackage{xcolor}         
\usepackage{amsmath}
\usepackage{amssymb}
\usepackage{mathtools}
\usepackage{amsthm}
\usepackage{multirow}
\usepackage{graphicx}
\usepackage{enumitem}
\usepackage{makecell}
\usepackage{float}
\usepackage{caption}
\usepackage{longtable}
\usepackage{wrapfig}
\usepackage{tabularx}
\usepackage{subcaption}
\usepackage{titletoc}
\title{MAGE: Towards Generalizable Multi-timescale\\EEG Representations}

\author{%
\parbox{\textwidth}{\centering
\textbf{Tuan-Kiet Doan}\textsuperscript{1} \quad
\textbf{Van-Chien Nguyen}\textsuperscript{1} \quad
\textbf{Trung-Hieu Tran}\textsuperscript{1} \quad
\textbf{Quang-Hung Pham}\textsuperscript{1}
\\[0.3em]
\textbf{Ngoc-Son Vu}\textsuperscript{2} \quad
\textbf{Huy Phan}\textsuperscript{3} \quad
\textbf{Phi Le Nguyen}\textsuperscript{4}
\\[0.3em]
\textbf{Duc-Han Le}\textsuperscript{1} \quad
\textbf{Nikola Simidjievski}\textsuperscript{1} \quad
\textbf{Van-Tam Nguyen}\textsuperscript{1}
\\[0.8em]
\normalfont\small
\textsuperscript{1}LTCI, Télécom Paris, Institut Polytechnique de Paris, France \\
\textsuperscript{2}LIST3N, Université de Technologie de Troyes, France \\
\textsuperscript{3}German Research Center for Artificial Intelligence (DFKI), Germany \\
\textsuperscript{4}Institute for AI Innovation and Societal Impact, Hanoi University of Science and Technology, Vietnam
\\[0.5em]
\texttt{\{tuan.doan,van-chien.nguyen,trung-hieu.tran,quang-hung.pham\}@ip-paris.fr} \\
\texttt{son.vu@utt.fr}, \texttt{huy.phan@dfki.de}, \texttt{lenp@soict.hust.edu.vn} \\
\texttt{\{duc-han.le,nikola.simidjievski,van-tam.nguyen\}@telecom-paris.fr}
}%
}

\begin{document}

\maketitle

\begin{abstract}
  Electroencephalography (EEG) data plays an important role in understanding human brain activity, with various major applications. While current approaches, which largely build on EEG foundation models, have demonstrated strong performance across datasets and tasks, they typically rely on a fixed temporal scale. This design choice limits their ability to capture characteristics of neural dynamic behavior, often spanning across multiple timescales, which are naturally captured in EEG recordings. To address this limitation, we propose MAGE (\textbf{Multi-timescale Adaptive Generalization for EEG}), a foundation model that dynamically adapts to capture and integrate information across different timescales. Specifically, MAGE combines two key mechanisms: (i) a novel multi-timescale attention mechanism, capable of learning representations across multiple temporal scales and (ii) a trainable aggregation module, that dynamically identifies and adapts to the most informative timescale of the input signal. Experimental results across seven datasets demonstrate that MAGE consistently outperforms existing EEG foundation models while using substantially fewer parameters. Moreover, our results support the hypothesis that different downstream tasks benefit from distinct temporal scales. The code is available at the \href{https://github.com/kietdoan/MAGE.git}{{\texttt{MAGE repository}}}.\looseness=-1
\end{abstract}

\section{Introduction}

One of science's biggest long-term goals is to understand and decode the human brain that drives applications in brain-computer interface~\cite{wolpaw2004control, wolpaw2013brain}, healthcare~\cite{ang2013brain, daly2008brain, koelstra2011deap}, and neuroscience~\cite{kamitani2005decoding, horikawa2013neural}. Electroencephalography (EEG) has emerged as a potential modality for this purpose, because of its accessibility and superior temporal resolution~\cite{alarcao2017emotions, flesher2021brain}. Various deep learning methods have been proposed for tasks like anomaly detection~\cite{roy2019chrononet, yildirim2020deep, van2019detecting}, sleep staging~\cite{zhao2022sleepcontextnet, phan2022automatic}, and motor imagery~\cite{hsu2023eeg, altaheri2023deep, hwaidi2022classification}. However, most of these approaches are typically task-specific and trained on small-scale datasets.

Recognizing these limitations, recent research has shifted toward foundation models~\cite{kostas2021bendr,jiang2024neurolm, zhang2023brant, cui2024neuro, ouahidi2025reve}. These models are able to overcome labeled data scarcity via self-supervised pre-training on large, unlabeled datasets.  While existing approaches focus on different aspects, such as spatio-temporal modeling~\cite{doner2025luna, yang2023biot} or noise-robust training~\cite{nguyen2026balanced, mohsenvand2020contrastive}, a common technique across these methods is the adoption of \textit{single-timescale tokenization}, which segments the continuous EEG signal into \textbf{fixed-size} windows. \looseness -1

Relying on a fixed window size creates a time-frequency bottleneck: a short window captures high-frequency components but loses low-frequency information, whereas a long window captures low-frequency oscillations but obscures transient details. Consequently, a single fixed scale limits the model's ability to capture the diverse frequency variations required by different tasks. For instance, motor imagery decoding relies on $\mu-$ and $\beta-$desynchronization \cite{pfurtscheller1999event, mcfarland2000mu, neuper2006erd}, sleep staging distinguishes $\delta$ and $\sigma$ waves~\cite{achermann1997low, berry2012aasm, steriade2006grouping}, while emotion recognition targets Frontal~$\alpha$ asymmetry and $\gamma$ oscillations~\cite{davidson1992anterior, muller1999processing, yang2020high}. Although recent works use multi-granularity fusion~\cite{wang2024adformer, wang2024medformer} or utilize foundation models with multi-kernel convolutions~\cite{zhou2025csbrain}, the former remain task-specific and limited by labeled data, even foundational approaches like~\cite{zhou2025csbrain} struggle to readily adapt to the relevant temporal resolution required by the task at hand. This raises a key research question: 

\textit{How do we enable multi-timescale learning, generalizable across different task and domains?}

To answer this question, we introduce \textbf{MAGE} (\textbf{M}ulti-timescale \textbf{A}daptive \textbf{G}eneralization for \textbf{E}EG). MAGE first generates multi-timescale tokens to capture spatiotemporal dynamics at each scale level. It then adaptively fuses these representations to produce a specific output, supporting both pretraining and downstream tasks. Specifically, we propose a \textit{Multi-timescale Tokenization} module to segment the raw signal using various patch sizes. Multi-timescale Tokenization enables the model to simultaneously capture fine-grained local patterns and coarse-grained global contexts, ensuring an effective representation of diverse brain activities. To encode these multi-timescale inputs, we propose a \textit{Multi-timescale Attention} module to enrich spatiotemporal representations. This module utilizes temporal attention to capture long-range dependencies within each timescale, while leveraging spatial attention to fuse information from spatially separated channels. To control the aggregation of these scales, we introduce an \textit{Aggregator} module. This module is designed to adaptively assign importance weights to each timescale level by analyzing the intrinsic temporal and spectral properties of the input to determine the importance of each scale. For example, sleep staging may favor coarse low-frequency patterns, whereas seizure detection often requires fine high-frequency transients. To the best of our knowledge, this is the first work to explore dynamic timescale aggregating for EEG analysis, establishing a new paradigm for enhancing the generalizability of EEG foundation models.

\noindent Our contributions are:
\begin{itemize}[topsep=0pt, itemsep=1pt, parsep=1pt]
    \item We introduce MAGE, a multi-timescale transformer-based foundation model for better generalized EEG decoding, which captures information temporally across different timescales and spatially across different channels.  
    \item We propose a novel Multi-timescale Attention module to extract rich spatiotemporal representations at each timescale, alongside an Aggregator that adaptively weights these scales based on input characteristics. \looseness -1
    \item The performance across seven EEG decoding tasks shows that MAGE outperforms existing foundation models across various tasks while using fewer parameters. We further analyze the contribution of each timescale, highlighting the necessity of multi-timescale decoding for EEG signals. 
\end{itemize}

\section{Related work}

\textbf{EEG Foundation Models. }Various works aim to build models that generate generalizable representations of EEG signals by training on a large volume of unlabeled data. To address signal noise, LaBram~\cite{jiang2024large} introduced a trainable neural tokenizer to generate discrete tokens, while EEGPT~\cite{wang2024eegpt} proposed a dual-loss pipeline combining semantic alignment with detailed reconstruction. Addressing the topological heterogeneity problem of EEG datasets, LuNa~\cite{doner2025luna} proposed a channel-unification module to map diverse channel sets into a unified space. CBraMod~\cite{wang2024cbramod} proposed a criss-cross attention mechanism, which attends to information across the temporal and spectral dimensions only. However, it is limited by a fixed timescale tokenization, which may have suboptimal representations for diverse downstream tasks. CSBrain~\cite{zhou2025csbrain} extended CBraMod's idea by employing cross-scale spatial-temporal tokenization using multiple kernel sizes. Nevertheless, while CSBrain captures features at different scales, it lacks a mechanism to dynamically prioritize the specific scale most relevant to each individual task. MAGE addresses this problem by adopting a trainable aggregator to dynamically decide the importance of each scale at each specific task.

\textbf{Multi-scale time series. }Multi-scale modeling has proved to be an effective approach to learn the temporal dynamic characteristics of time series signals. Architectures like TimeMixer~\cite{wang2024timemixer} and TimeMixer++~\cite{wang2024timemixer++} adopt a sequence of pooling layers to downsample the input, generating a multi-scale representation that captures both global trends and local seasonal patterns. Meanwhile, PathFormer~\cite{chen2024pathformer}, MPTSNet~\cite{mu2025mptsnet}, and AdaMixT~\cite{zhang2025adamixt} segment the series into varying timescales and aggregate information from these scales within each Transformer block. However, these methods are designed for general time-series forecasting tasks, which involve signals with lower sampling rates and higher signal-to-noise ratios than EEG. While several approaches~\cite{wang2024medformer, wang2024adformer, li2025mhanet, feng2026msanet, fan2025dmf2mel} have used multi-timescale modeling of medical signals, they have been designed as a task-specific architecture, limiting their applicability as a universal foundation model. Consequently, the application of multi-scale segmentation for high-frequency, noisy EEG analysis remains largely unexplored. MAGE overcomes many of the limitations of these methods, which typically rely on task-specific training pipelines. By leveraging a pre-training strategy, MAGE dynamically aggregates information across diverse timescales during downstream fine-tuning, successfully circumventing the feature suppression limitations inherent in existing fixed-scale methods.

\textbf{Mixture of Experts (MoE). }MoE is a network that dynamically activates different submodules, or experts, dependent on the input~\cite{mu2025comprehensive}. Recently, MoE has attracted much attention to addressing the difficulties of training in diverse and conflicting tasks~\cite{liu2024deepseek}. While MoE is traditionally adopted to route data across different subnetworks, MAGE adapts this paradigm to the temporal domain. In our framework, each timescale is treated as an individual expert, with an aggregator dynamically fusing multi-scale information based on the signal’s intrinsic properties.\looseness -1

\begin{figure*}
    \centering
    \includegraphics[width=0.95
    \textwidth]{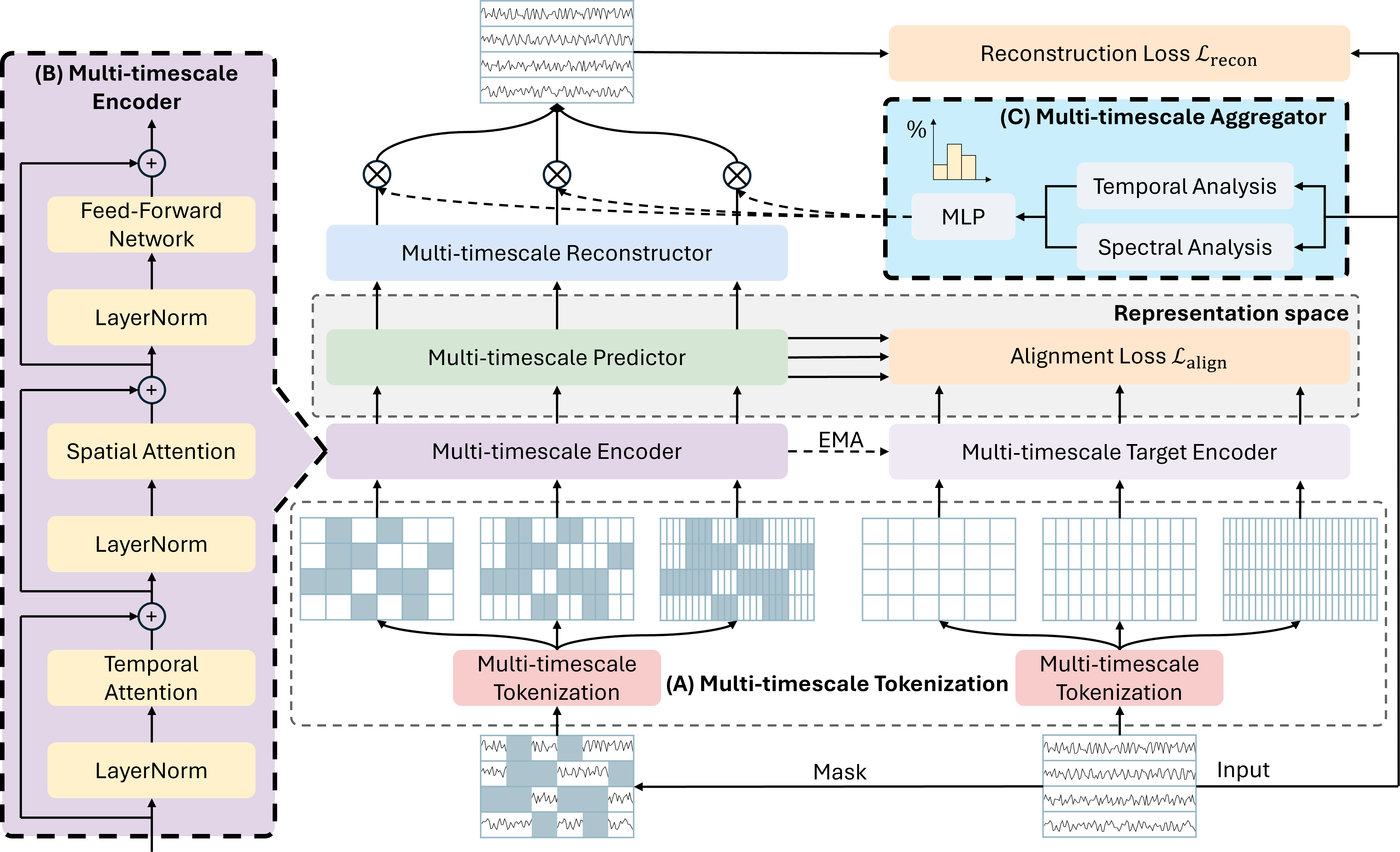}
    
    \caption{Overview of the MAGE pretraining pipeline. MAGE employs (A) Multi-timescale tokenization and (B) Multi-timescale encoder, optimized with a dual-loss scheme for both semantic and fine-grained learning. (C) Multi-timescale Aggregator dynamically weights each scale for the final reconstruction. }
    \label{fig:MoG_these}
\end{figure*}

\section{MAGE}

Figure~\ref{fig:MoG_these} presents an overview of MAGE, our proposed multi-timescale foundation model for EEG representation. MAGE is composed of three main stages: (i) Multi-timescale Tokenization, which extracts features at different timescales (Sec.~\ref{subsec: multi-gran token}); (ii) Multi-timescale Attention, which captures complex spatiotemporal dependencies in EEG signals (Sec.~\ref{subsec: multi-scale attn}); and (iii) Multi-timescale Aggregator, which dynamically weights each timescale based on input-dependent characteristics (Sec.~\ref{subsec: aggregator}). The full training pipeline is described in Sec.~\ref{subsec: train with MAGE}.

\subsection{Multi-timescale Tokenization}\label{subsec: multi-gran token}

Let $\mathbf{E}\in \mathbb{R}^{C\times T}$ denote EEG signals, with $C$ channels and $T$ timestamps. Due to the complex characteristics of EEG signals, different decoding tasks will depend on different frequencies of interest. However, previous EEG foundation models use fixed patch sizes, limiting generalizability across different decoding tasks. We, therefore, introduce Multi-timescale Tokenization, which segments $\mathbf{E}$ into tokens at multiple patch sizes, capturing both high-frequency transients and low-frequency long-range patterns for a multiscale representation. Let $\mathcal{P} = \{ p_1, \dots, p_G \}$ denote the set of $G$ patch sizes. For each $p_g \in \mathcal{P}$, $\mathbf{E}$ is segmented into non-overlapping patches of length $p_g$, resulting in $\mathbf{E}_{p_g} \in \mathbb{R}^{C \times n_g \times p_g}$ with $n_g = \lfloor T / p_g \rfloor$ segments per channel.

To effectively capture both local temporal and frequency information within each segment, each patch $\mathbf{e}_{p_g} \in \mathbb{R}^{p_g}$ is processed by a dual-branch spectral-temporal embedding module~\cite{wang2024cbramod}. The \textbf{temporal branch} extracts local temporal features via 1D convolutions, while the \textbf{spectral branch} captures frequency information via FFT followed by a linear transformation:
\begin{equation}
    \mathbf{x}_{p_g} = \phi_{\text{temp}}(\mathbf{e}_{p_g})) + \phi_{\text{spec}}(|\text{FFT}(\mathbf{e}_{p_g})|) \in \mathbb{R}^D,
\end{equation}
where $D$ is the feature dimension. After the tokenization module, the raw EEG signal $\mathbf{E}$ is transformed to a set of final embeddings $\mathcal{X}_\mathcal{P} = \{ \mathbf{X}^{(g)} \}_{g=1}^G$, where $\mathbf{X}^{(g)} \in \mathbb{R}^{C \times n_g \times D}$.

\subsection{Multi-timescale Attention}\label{subsec: multi-scale attn}
Multivariate EEG signals exhibit complex spatiotemporal dynamics, involving both temporal dynamics within each channel and spatial connectivity across them. We decompose representation learning into two stages: the first captures temporal dynamics within each channel, and the second models spatial interactions across channels. This design allows the model to focus attention on the most relevant spatiotemporal regions. We propose the Multi-timescale Attention (MtA) module to process the multi-timescale input $\mathcal{X}_\mathcal{P} = \{ \mathbf{X}^{(g)} \}_{g=1}^G$. MtA consists of two orthogonal attention modules (Temporal Attention module and Spatial Attention module) followed by a Feed-forward Network (Fig.~\ref{fig:MoG_these}).

\textbf{Temporal Attention module } captures sequential dynamics within each channel. For timescale $\mathbf{X}^{(g)}$ and channel $c\in\{1,\dots,C\}$, the input tensor is sliced into the temporal sequence $\mathbf{X}_{c} \in \mathbb{R}^{n_g \times D}$.

Query, key, and value projections are computed as $\mathbf{Q}_c = \mathbf{X}_{c}\mathbf{W}^T_Q$, $\mathbf{K}_c = \mathbf{X}_{c}\mathbf{W}^T_K$, and $\mathbf{V}_c = \mathbf{X}_{c}\mathbf{W}^T_V$. Rotary Positional Encoding $\mathcal{R}(\cdot)$ is used to preserve
temporal order. The temporal attention output $\mathbf{F}_{c} \in \mathbb{R}^{n_g \times D}$ is obtained via the output projection $\mathbf{W}^T_O$:
\begin{equation}
    \mathbf{F}_{c} = \left[\operatorname{softmax}\left( \frac{\mathcal{R}(\mathbf{Q}_c) \mathcal{R}(\mathbf{K}_c)^\top}{\sqrt{D}} \right) \mathbf{V}_c\right]\mathbf{W}^T_O + \mathbf{X}_{c}.
\end{equation}

Outputs from all channels are then concatenated to form the intermediate feature tensor $\mathbf{F}^{(g)} \in \mathbb{R}^{C \times n_g \times D}$.
\textbf{Spatial Attention module} captures dependencies across EEG channels. This attention mechanism is applied sequentially after the temporal attention module. To preserve spatial identity, a learnable channel embedding is added to each channel's representation as inspired by \cite{wang2024eegpt, yang2026eeg}. For time step $t \in \{1,\dots,n_g\}$, let $\mathbf{F}_{t} \in \mathbb{R}^{C \times D}$ denote features across channels. Spatial projections are $\mathbf{Q}'_t = \mathbf{F}_{t}\mathbf{W}^S_Q$, $\mathbf{K}'_t = \mathbf{F}_{t}\mathbf{W}^S_K$, and $\mathbf{V}'_t = \mathbf{F}_{t}\mathbf{W}^S_V$. The spatial interaction is then computed as:
\begin{equation}
    \mathbf{U}_{t} = \left[\operatorname{softmax}\left( \frac{\mathbf{Q}'_t (\mathbf{K}'_t)^\top}{\sqrt{D}} \right) \mathbf{V}'_t\right]\mathbf{W}^S_O + \mathbf{F}_{t}.
\end{equation}
 
\textbf{Feed-forward Network (FFN) } integrates the features via two linear layers with a GeLU activation. The final representation after the MtA module is:
$\mathbf{Y}^{(g)}=\mathbf{U}^{(g)} + \mathrm{FFN}\!\left(\mathbf{U}^{(g)}\right).$

\subsection{Multi-timescale Aggregator}\label{subsec: aggregator}

MtA enables multi-timescale modeling, but distinct EEG decoding tasks rely on specific frequency bands inherent to their temporal characteristics. Increasing timescales naively may add redundancy or noise rather than informative features, and manual tuning the scale is time-consuming. We, therefore, introduce a lightweight Aggregator module to dynamically weight each timescale for better generalization across tasks. Given raw EEG $\mathbf{E} \in \mathbb{R}^{C \times T}$, the Aggregator uses two parallel temporal and frequency branches to determine the importance of each timescale.

\textbf{Temporal Analysis. }We employ a 1D convolutional network to encode high-level semantic features from the raw sequence. To have a global analysis, a global average pooling layer is adopted to compresses the temporal dimension to capture the overall distribution of features, resulting in a global embedding $\mathbf{a}_{\text{temp}}$ that is robust to temporal shifts.

\textbf{Frequency Analysis. }To capture global periodicities and seasonal information, the signal is transformed via FFT. We process the Magnitude and Phase components separately with two convolutional modules. The resulting representations are concatenated to produce the comprehensive frequency embedding $\mathbf{a}_{\text{freq}}$.

The final decision head, a multi-layer perceptron, concatenates the embeddings $\mathbf{a}_{\text{temp}}$ and $\mathbf{a}_{\text{freq}}$ to form $\mathbf{a}$, and outputs dynamic weight for each timescale: 
\begin{equation}
    \boldsymbol{\alpha} = \operatorname{softmax}(\mathrm{MLP}(\mathbf{a})) \in \mathbb{R}^G.
\end{equation} 
\subsection{Training with MAGE}\label{subsec: train with MAGE}
\subsubsection{Pretraining pipeline}
To extract both semantic and fine-grained information, we extend the EEGPT framework~\cite{wang2024eegpt, nguyen2026balanced} for multi-timescale training with alignment and reconstruction losses. The raw signal $\mathbf{E}$ is first processed by a Multi-timescale Tokenization block to produce token sequences $\mathcal{X}_\mathcal{P} = \{ \mathbf{X}^{(g)} \}_{g=1}^G$.

\textbf{Alignment loss. }
We use a teacher-student architecture to learn robust semantic representations. Let $\mathcal{M} = \{ \mathbf{M}^{(g)} \}_{g=1}^G$, here $\mathbf{M}^{(g)}$ denote binary masks for tokens $\mathbf{X}^{(g)}$. 

The student network processes masked input $\tilde{\mathcal{X}}_\mathcal{P} = \{ \mathbf{X}^{(g)} \odot \mathbf{M}^{(g)} \}_{g=1}^G$. The teacher network has the identical architecture with the student version, and its parameters are updated via Exponential Moving Average (EMA) of student parameters. This module processes unmasked input $\mathcal{X}_\mathcal{P}$ to produce stable latent representations:
\begin{equation}
    \mathcal{Z}_S = \mathrm{Enc}_S(\tilde{\mathcal{X}}_\mathcal{P}), \mathcal{H}_T = \{ \mathbf{H}_T^{(g)} \}_{g=1}^G = \mathrm{Enc}_T(\mathcal{X}_\mathcal{P}),
\end{equation}
where the indices $S$ and $T$ refer to the student and teacher, respectively. The Predictor maps visible tokens $\mathcal{Z}_S$ to $\hat{\mathcal{H}}_S= \{ \mathbf{\hat{H}}_S^{(g)} \}_{g=1}^G = \mathrm{Pred}\left( \mathcal{Z}_S \right)$. Both Encoder and Predictor are stacks of MtA modules. The alignment loss is computed as:
\begin{equation}
    \mathcal{L}_{\text{align}} = \displaystyle \sum_{g=1}^G\left( \| \hat{\mathbf{H}}^{(g)}_S - \mathbf{H}^{(g)}_T \| \right).
\end{equation}
\textbf{Reconstruction loss. } To reconstruct the original signal, latent outputs from all timescales are aggregated. While coarse timescale captures low-frequency components, fine timescale captures high-frequency features. The aggregator is adopted to dynamically weights each timescale. A key challenge when training with aggregator is mode collapse, akin to the ``expert collapse'' in MoE~\cite{liu2024deepseek, shazeer2017outrageously}. The aggregator may collapse by assigning nearly all weight to one timescale, blocking updates to others. We mitigate this by injecting Gaussian noise into the logits before normalization, ensuring gradients are distributed across all timescales.

Formally, stochastic importance weights $\tilde{\boldsymbol{\alpha}}$ are computed by perturbing the decision boundary:
\begin{equation}
    \tilde{\boldsymbol{\alpha}} = \operatorname{softmax}\left( \mathrm{MLP}(\mathbf{a}) + \boldsymbol{\epsilon} \right), \text{where } \boldsymbol{\epsilon} \sim \mathcal{N}(0, \sigma^2 \mathbf{I}).
\end{equation}
The reconstructed signal $\mathbf{\hat{E}}$ is:
\begin{equation}
    \mathbf{\hat{E}} = \sum_{g=1}^G \tilde{\alpha}_g \cdot \mathrm{Dec}^{(g)}\left( \mathbf{\hat{h}}^{(g)}_S \right),
\end{equation}
where $\mathrm{Dec}^{(g)}()$ is a linear decoder of scale $g$. Reconstruction error is minimized via $\mathcal{L}_{\text{recon}}=\|\mathbf{E}-\mathbf{\hat{E}}\|_2^2$. The final loss is: 
\begin{equation}
    \mathcal{L} = \lambda_{\text{recon}} \mathcal{L}_{\text{recon}} + \lambda_{\text{align}} \mathcal{L}_{\text{align}},
\end{equation}
where $\lambda_{\text{recon}}$ and $\lambda_{\text{align}}$ balance the two tasks.

\subsubsection{Downstream task adaptation}
Similarly to pre-training, during downstream, stochastic noise is added to avoid reliance on a single timescale. The pre-trained encoder is frozen during downstream adaptation to reduce overfitting. The task prediction $\hat{\mathbf{y}}$ is obtained by aggregating the frozen representations via a scale-specific head:
\begin{equation}
    \hat{\mathbf{y}} = \sum_{g=1}^G \tilde{\alpha}_g \cdot \mathrm{Classifier}^{(g)}\left( \mathbf{Z}^{(g)}_S \right),
\end{equation}
where $\mathrm{Classifier}^{(g)}$ is the task-specific head for timescale $g$, and $\tilde{\boldsymbol{\alpha}}$ is the stochastic importance vector derived from the aggregator. Notably, this stochasticity is disabled during evaluation to ensure deterministic predictions. See Appendix~\ref{appendix: downstream} for details.

\section{Experimental results}
\subsection{Pretraining setup}
\textbf{Data processing. }We pretrain MAGE on a subset of the Temple University Hospital EEG corpus (TUEG) dataset, with around 500 hours of training data. We opted for this subset size because our preliminary experiments scaling up to 2,500 hours (detailed in the Appendix~\ref{subsec: pretraining_hour}) demonstrated marginal improvements, suggesting diminishing returns at larger data scales. We follow the prior preprocessing steps~\cite{wang2024cbramod, zhou2025csbrain}, detailed is in Appendix~\ref{appendix: pretrain}. To avoid amplitude mismatches across devices, we apply z-score normalization per channel to each sample. 

\textbf{Pretraining setting. }Following prior work~\cite{wang2024cbramod, zhou2025csbrain}, we select 19 standard channels from 10-20 international system to pretrain MAGE. We select three different patch sizes: 50, 100, and 200, corresponding to 250ms, 500ms and 1s, respectively. During pretraining, we randomly mask $70\%$ of the EEG samples and adopt the AdamW optimizer~\cite{loshchilov2017decoupled}. For the multi-timescale aggregator, a noise with zero mean and 0.3 standard deviation is applied.
\looseness-1

\subsection{Downstream task setup}
\textbf{Evaluation datasets. }We conduct experiments on seven publicly available EEG datasets to evaluate the generalizability of our model. These datasets include BCIC-IV-2a~\cite{brunner2008bci}, PhysioNet-MI~\cite{goldberger2000physiobank, schalk2004bci2000}, SHU-MI~\cite{ma2022large}, FACED~\cite{chen2023large}, TUAB~\cite{obeid2016temple}, KaggleERN~\cite{margaux2012objective}, and Sleep-EDFx~\cite{kemp2000analysis}. Unlike previous work that uses a fixed split of training, evaluation, and testing sets, in each data set, we conduct experiments many times, with a different set of training, evaluation, and testing subjects. We then report the mean and standard deviation across these different sets to summarize the overall performance. Further details are provided in the Appendix ~\ref{appendix: dataset}. 

\textbf{Baselines. }To evaluate the effectiveness of MAGE, we compare it with several current state-of-the-art foundation models for EEG, which include LaBram~\cite{jiang2024large}, BIOT~\cite{yang2023biot}, EEGPT~\cite{wang2024eegpt}, CBraMod~\cite{wang2024cbramod}, and CSBrain~\cite{zhou2025csbrain}. Following standard protocol, we keep the pre-trained encoder frozen across all experiments and append a consistent classifier head to assess the quality of the learned representations. Additionally, we compare MAGE with Medformer~\cite{wang2024medformer}, which is a supervised multi-scale model for medical time-series classification. Following prior work, we use Balanced Accuracy, Cohen's Kappa, and Weighted F1 for multi-class classification problems, while we use Balanced Accuracy, AUC-PR, and AUROC for binary classification. More details about these baselines and metrics are illustrated in the Appendix~\ref{appendix: model} and Appendix~\ref{appendix: metrics}.

\textbf{Statistical significance test. }The results for EEG signals often have large standard deviation. This is mainly because the substantial inter-subject variability in EEG signals. In our evaluation protocol, each fold uses a different test set, resulting in considerable variation in performance across folds. Therefore, aggregating results over these highly variable folds produces large standard deviations that may obscure the true performance differences. To provide further evidence of performance, we employed a corrected two-sided paired t-test ($p < 0.05$)~\cite{nadeau1999inference} across all folds and 3 random seeds. The corrected paired t-test is designed to account for the non-independence of cross-validation folds. Demonstrating significance under this strict condition confirms the high reliability of the method. \looseness-1

\subsection{Experimental Results}
\begin{table*}[tbp]
\caption{Comparison of MAGE with existing foundation models across various downstream tasks, metrics, and test subjects. Numbers under model names indicate parameter sizes in millions (M). MAGE-froz denotes MAGE where the aggregator is kept frozen during downstream evaluation. MAGE achieves the highest average ranking among all models. \textbf{Bold} indicates the best result, and \underline{underlined} indicates the second best. The symbol $\downarrow$ indicates that the baseline model is significantly worse than MAGE, while $\uparrow$ indicates that the baseline model is significantly better than MAGE based on the corrected two-sided paired t-test ($p < 0.05$). The average rank is calculated across all metrics.}
\label{tab:performance_after_rebuttal}
\centering
\resizebox{\textwidth}{!}{%
    \begin{tabular}{llcccccccc}
    \toprule
    \multirow{2}{*}{Dataset} &
    \multirow{2}{*}{Metrics} & Medformer & LaBram & BIOT & EEGPT & CBraMod & CSBrain & MAGE & MAGE-froz \\
    &   & 1.4M & 5.8M & 3.2M &25.0M & 4.0M & 8.9M & \textbf{0.6M} & \textbf{0.6M} \\
    \midrule
    \multirow{3}{*}{BCIC-IV-2a} 
    & bAcc   & 52.68$_{{\pm 6.06}}$$\downarrow$ & 49.75$_{{\pm 8.73}}$$\downarrow$ & 49.92$_{{\pm 11.74}}$& 50.49$_{{\pm 6.38}}$$\downarrow$ & 46.09$_{{\pm 6.63}}$$\downarrow$ & 55.11$_{{\pm 8.63}}$ & \underline{56.78}$_{{\pm 6.08}}$ & \textbf{60.81}$_{{\pm 10.05}}$ \\
    & Kappa  & 36.91$_{{\pm 8.08}}$$\downarrow$ & 33.00$_{{\pm 11.64}}$$\downarrow$ & 33.23$_{{\pm 15.65}}$ & 33.98$_{{\pm 8.50}}$$\downarrow$ & 28.12$_{{\pm 8.83}}$$\downarrow$ & 40.14$_{{\pm 11.50}}$ & \underline{42.38}$_{{\pm 8.11}}$ & \textbf{47.75}$_{{\pm 13.40}}$ \\
    & wF1    & 52.08$_{{\pm 6.34}}$$\downarrow$ & 47.64$_{{\pm 10.12}}$$\downarrow$ & 48.06$_{{\pm 13.36}}$ & 49.61$_{{\pm 6.77}}$$\downarrow$ & 45.02$_{{\pm 6.96}}$$\downarrow$ & 54.12$_{{\pm 8.96}}$ & \underline{55.61}$_{{\pm 6.60}}$ & \textbf{59.30}$_{{\pm 11.72}}$ \\
    \midrule
    \multirow{3}{*}{PhysioNet} 
    & bAcc   & 59.01$_{{\pm 3.39}}$$\downarrow$ & 58.82$_{{\pm 4.40}}$$\downarrow$ & 55.02$_{{\pm 3.33}}$$\downarrow$ & 53.80$_{{\pm 2.91}}$$\downarrow$ & 52.22$_{{\pm 4.01}}$$\downarrow$ & 60.62$_{{\pm 3.14}}$$\downarrow$ & \underline{62.69}$_{{\pm 3.59}}$ & \textbf{64.08}$_{{\pm 3.51}}$ \\
    & Kappa  & 45.36$_{{\pm 4.52}}$$\downarrow$ & 45.11$_{{\pm 5.86}}$$\downarrow$ & 40.05$_{{\pm 4.43}}$$\downarrow$ & 38.41$_{{\pm 3.88}}$$\downarrow$ & 36.30$_{{\pm 5.35}}$$\downarrow$ & 47.52$_{{\pm 4.18}}$$\downarrow$ & \underline{50.27}$_{{\pm 4.77}}$ & \textbf{52.12}$_{{\pm 4.66}}$ \\
    & wF1    & 58.92$_{{\pm 3.39}}$$\downarrow$ & 58.86$_{{\pm 4.38}}$$\downarrow$ & 55.10$_{{\pm 3.28}}$$\downarrow$ & 53.78$_{{\pm 2.89}}$$\downarrow$ & 52.20$_{{\pm 4.04}}$$\downarrow$ & 60.54$_{{\pm 3.17}}$$\downarrow$ & \underline{62.71}$_{{\pm 3.54}}$ & \textbf{64.19}$_{{\pm 3.51}}$ \\
    \midrule
    \multirow{3}{*}{FACED}      
    & bAcc   & 40.61$_{{\pm 1.40}}$$\downarrow$ & 46.20$_{{\pm 1.80}}$$\downarrow$ & 45.04$_{{\pm 2.13}}$$\downarrow$ & 44.12$_{{\pm 1.80}}$$\downarrow$ & 39.88$_{{\pm 2.06}}$$\downarrow$ & 46.80$_{{\pm 1.59}}$$\downarrow$ & \underline{51.42}$_{{\pm 2.11}}$ & \textbf{51.44}$_{{\pm 2.21}}$ \\
    & Kappa  & 32.92$_{{\pm 1.51}}$$\downarrow$ & 39.24$_{{\pm 1.98}}$$\downarrow$ & 37.92$_{{\pm 2.35}}$$\downarrow$ & 36.85$_{{\pm 2.05}}$$\downarrow$ & 32.11$_{{\pm 2.18}}$$\downarrow$ & 39.81$_{{\pm 1.74}}$$\downarrow$ & \underline{44.97}$_{{\pm 2.31}}$ & \textbf{44.98}$_{{\pm 2.56}}$ \\
    & wF1    & 40.25$_{{\pm 1.30}}$$\downarrow$ & 46.04$_{{\pm 1.74}}$$\downarrow$ & 44.77$_{{\pm 1.94}}$$\downarrow$ & 43.81$_{{\pm 1.79}}$$\downarrow$ & 39.61$_{{\pm 1.91}}$$\downarrow$ & \underline{46.32}$_{{\pm 1.48}}$$\downarrow$ & \textbf{50.94}$_{{\pm 1.97}}$ & \textbf{50.94}$_{{\pm 2.33}}$ \\
    \midrule
    \multirow{3}{*}{Sleep-EDFx} 
    & bAcc   & 61.92$_{{\pm 5.75}}$$\downarrow$ & 62.89$_{{\pm 3.72}}$$\downarrow$ & \textbf{70.64}$_{{\pm 3.43}}$$\uparrow$ & 66.54$_{{\pm 4.00}}$$\downarrow$ & \underline{70.63}$_{{\pm 4.11}}$ & 70.02$_{{\pm 3.78}}$ & 70.08$_{{\pm 4.02}}$ & 66.36$_{{\pm 3.21}}$ \\
    & Kappa  & 58.43$_{{\pm 7.58}}$$\downarrow$ & 63.96$_{{\pm 5.22}}$$\downarrow$ & 68.12$_{{\pm 5.39}}$ & 65.46$_{{\pm 5.96}}$$\downarrow$ & \underline{68.79}$_{{\pm 6.41}}$ & \textbf{68.89}$_{{\pm 5.56}}$ & \textbf{68.89}$_{{\pm 5.98}}$ & 65.86$_{{\pm 5.70}}$ \\
    & wF1    & 68.69$_{{\pm 5.87}}$$\downarrow$ & 73.30$_{{\pm 4.34}}$$\downarrow$ & 77.02$_{{\pm 3.96}}$ & 74.60$_{{\pm 4.65}}$$\downarrow$ & \textbf{77.65}$_{{\pm 4.36}}$ & \underline{77.45}$_{{\pm 4.18}}$ & 77.33$_{{\pm 4.47}}$ & 74.74$_{{\pm 3.92}}$ \\
    \midrule
    \multirow{3}{*}{TUAB}       
    & bAcc   & 78.72$_{{\pm 0.53}}$ & 78.61$_{{\pm 0.35}}$ & \underline{79.15}$_{{\pm 0.97}}$ & 78.32$_{{\pm 0.34}}$$\downarrow$ & 77.08$_{{\pm 0.85}}$$\downarrow$ & 77.79$_{{\pm 0.58}}$$\downarrow$ & \textbf{79.79}$_{{\pm 0.89}}$ & 79.01$_{{\pm 0.58}}$ \\
    & PR     & 84.40$_{{\pm 0.91}}$$\downarrow$ & \underline{86.83}$_{{\pm 0.24}}$ & 86.21$_{{\pm 1.59}}$ & 85.51$_{{\pm 0.94}}$$\downarrow$ & 84.93$_{{\pm 1.02}}$$\downarrow$ & 86.17$_{{\pm 0.62}}$ & \textbf{86.85}$_{{\pm 0.88}}$ & 86.29$_{{\pm 1.07}}$ \\
    & ROC    & 86.29$_{{\pm 0.44}}$ & \underline{86.90}$_{{\pm 0.29}}$ & 86.61$_{{\pm 0.88}}$ & 85.94$_{{\pm 0.39}}$$\downarrow$ & 85.82$_{{\pm 0.83}}$ & 86.39$_{{\pm 0.58}}$ & \textbf{87.13}$_{{\pm 0.80}}$ & 86.76$_{{\pm 0.44}}$ \\
    \midrule
    \multirow{3}{*}{KaggleERN}  
    & bAcc   & 59.76$_{{\pm 1.17}}$ & 56.15$_{{\pm 1.46}}$$\downarrow$ & 54.18$_{{\pm 1.65}}$$\downarrow$ & 57.32$_{{\pm 1.59}}$$\downarrow$ & 56.38$_{{\pm 1.26}}$$\downarrow$ & 57.56$_{{\pm 1.39}}$$\downarrow$ & \underline{60.92}$_{{\pm 1.32}}$ & \textbf{61.10}$_{{\pm 2.14}}$ \\
    & PR     & 77.77$_{{\pm 1.29}}$ & 77.12$_{{\pm 0.67}}$ & 75.05$_{{\pm 1.97}}$$\downarrow$ & 77.17$_{{\pm 1.85}}$& 76.96$_{{\pm 1.19}}$& 77.65$_{{\pm 2.22}}$ & \underline{78.40}$_{{\pm 1.91}}$ & \textbf{79.23}$_{{\pm 2.05}}$ \\
    & ROC    & 63.29$_{{\pm 1.32}}$ & 60.86$_{{\pm 1.04}}$$\downarrow$ & 57.25$_{{\pm 2.20}}$$\downarrow$ & 61.19$_{{\pm 2.47}}$$\downarrow$ & 60.18$_{{\pm 1.34}}$$\downarrow$ & 61.88$_{{\pm 2.55}}$ & \underline{64.68}$_{{\pm 1.97}}$ & \textbf{65.58}$_{{\pm 2.30}}$ \\
    \midrule
    \multirow{3}{*}{SHU-MI}     
    & bAcc   & 61.42$_{{\pm 3.26}}$ & \textbf{63.27}$_{{\pm 3.65}}$ & 60.55$_{{\pm 3.25}}$ & 60.01$_{{\pm 2.80}}$$\downarrow$ & 62.40$_{{\pm 3.68}}$ & \underline{62.56}$_{{\pm 3.67}}$ & 62.22$_{{\pm 3.49}}$ & 61.87$_{{\pm 2.56}}$ \\
    & PR     & 66.67$_{{\pm 4.71}}$ & \textbf{69.34}$_{{\pm 5.43}}$ & 65.86$_{{\pm 4.72}}$ & 63.53$_{{\pm 3.51}}$$\downarrow$ & 68.05$_{{\pm 5.95}}$ & \underline{69.03}$_{{\pm 5.34}}$ & 68.63$_{{\pm 4.77}}$ & 68.24$_{{\pm 4.00}}$ \\
    & ROC    & 66.59$_{{\pm 4.52}}$ & \textbf{69.23}$_{{\pm 5.21}}$ & 65.73$_{{\pm 4.73}}$ & 64.05$_{{\pm 3.72}}$$\downarrow$ & 68.10$_{{\pm 5.25}}$ & \underline{68.80}$_{{\pm 4.66}}$ & 67.91$_{{\pm 4.73}}$ & 67.66$_{{\pm 3.80}}$ \\
    \midrule
    \multicolumn{2}{l}{\textbf{Average Rank}} & 5.43 & 4.76 & 5.48 & 6.14 & 6.29 & 3.31 & \textbf{2.14} & \underline{2.45} \\
    \bottomrule
    \end{tabular}%
}
\end{table*}
\textbf{MAGE validates the superiority of multi-timescale method over rigid fixed-patch baselines. }As shown in Table~\ref{tab:performance_after_rebuttal}, MAGE achieves a better or comparable performance compared to other models across various tasks. In BCIC-IV-2a, PhysioNet-MI, FACED, TUAB, and KaggleERN, MAGE outperforms the other models with a gap of $2-3\%$. This consistent gap highlights MAGE's superiority, which stems from its use of multiple patch sizes to effectively capture diverse temporal dynamics. On the Sleep-EDFx and SHU-MI datasets, MAGE performs comparably to the top model, CBraMod. Overall, MAGE earns the best average rank of 2.14. It easily surpasses the runner-up, CSBrain, which ranks at 3.31. These results provide strong empirical evidence for our core hypothesis: effective decoding requires adapting temporal granularity to the specific task. Impressively, MAGE achieves these state-of-the-art results with only 0.6M parameters, a fraction of the size of contemporary foundation models. This demonstrates that dividing signals into multi-timescale tokens significantly enhances modeling performance. When we conduct a statistical significance test, despite the large variance across different evaluation subjects, MAGE still performs significantly better than all baselines on PhysioNet-MI and FACED. On BCIC2a, KaggleERN, and Sleep-EDFx, it significantly outperforms almost other foundation models while for TUAB and SHU-MI, MAGE achieves comparable results across the board. Notably, MAGE demonstrates no statistically significant drop in performance compared to any baseline across all evaluated datasets, except when compared to BIOT on Sleep-EDFx using the Balanced Accuracy metric.

\textbf{Training the Aggregator at each downstream task yields better performance. }Table~\ref{tab:performance_after_rebuttal} also compares the effects of freezing versus fine-tuning the aggregator during downstream tasks. Although a frozen aggregator remains reasonably effective on datasets where individual scales inherently perform well, such as PhysioNet-MI or BCIC-IV-2a, it struggles when there is a significant performance disparity between scales. For instance, on the Sleep-EDFx dataset, freezing the aggregator results in a $3-4\%$ performance drop, highlighting the need to fine-tune the aggregator to properly balance diverse scales. While a frozen aggregator already leads to competitive performance, the ability to adaptively adjust weights via a trainable aggregator is a significant structural advantage of MAGE over other baselines. In practical applications, this adaptability can be particularly beneficial, as the optimal aggregation weights for downstream tasks do not necessarily remain the same as those learned during pre-training. \looseness -1

\begin{wrapfigure}{r}{0.6\textwidth}
    \vspace{2pt}
    \centering
    \includegraphics[width=\linewidth]{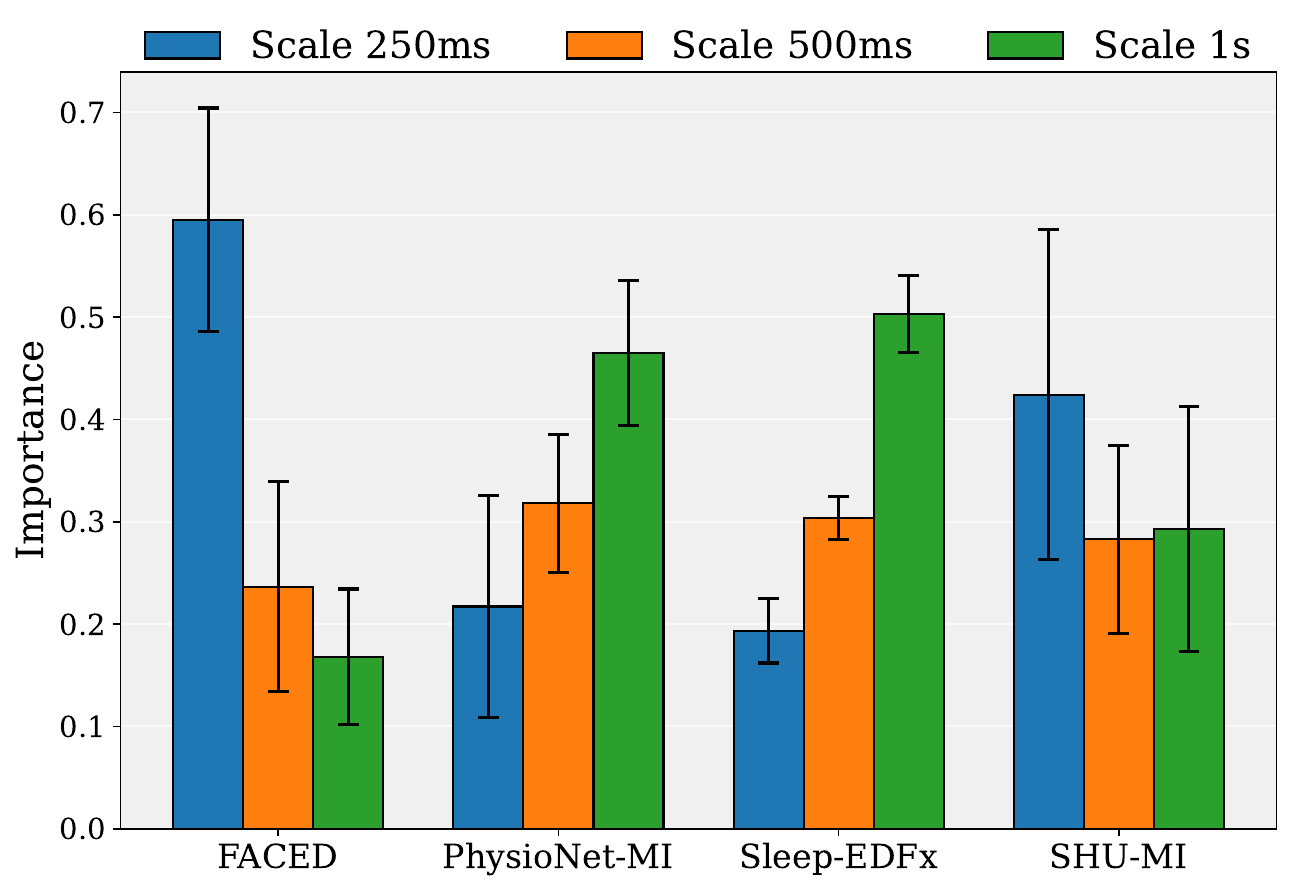}
    \caption{Distribution of importance weights across timescales by dataset. It can be shown that each dataset requires a different scale to capture relevant information.}
    \label{fig:scale_importance}
    \vspace{-0.1in}
\end{wrapfigure}

\textbf{The importance of each timescale is task-dependent.} Figure~\ref{fig:scale_importance} demonstrates the weight distribution of three timescales across four datasets: FACED, PhysioNet-MI, Sleep-EDFx, and SHU-MI. The results indicate that FACED and SHU-MI rely heavily on the finest timescale (250ms), assigning it weights of $60\%$ and $45\%$ respectively, whereas the coarse timescale (1s) makes a small contribution. Meanwhile, the Sleep-EDFx and PhysioNet-MI prioritize larger patch sizes, evidenced by more than $45\%$ weight for the scale 1s compared to less than $20\%$ for the scale 250ms. These findings suggest that while Sleep-EDFx necessitates long-window contexts, emotion recognition like FACED benefit more from short-patch information, providing valuable insight for selecting task-specific patch sizes. Notably, we observe that the aggregator adapts dynamically to each individual sample rather than outputting a nearly fixed value for each dataset. More detail about how the aggregator adapts to each individual sample is illustrated in the Appendix.~\ref{appendix: sample level analysis}.

\begin{table*}
\caption{Comparison of different timescales performance. MAGE achieves a better performance than any single timescales, while MAGE-router is able to achieve a fair comparison with the best single scale in some tasks. \textbf{Bold} indicates the best result, and \underline{underlined} indicates the second best. The average rank is calculated across all metrics. The result ablation is reported on 1 seed.}
\label{tab:single_granularity}
\centering

\resizebox{0.95\textwidth}{!}{%
    \begin{tabular}{llcccccc}
    \toprule
    Dataset & Metrics & 250ms only & 500ms only & 1s only & MAGE-router & MAGE\\
    \midrule
    \multirow{3}{*}{FACED} & Balanced Acc & \underline{50.93$_{\pm 2.82}$} & 49.78$_{\pm 1.80}$ & 46.65$_{\pm 1.84}$ & 49.87$_{\pm 2.45}$ & \textbf{51.30$_{\pm 3.11}$} \\
                                & Cohen's Kappa  & \underline{44.41$_{\pm 3.19}$} & 43.14$_{\pm 1.97}$ & 39.79$_{\pm 1.98}$ & 43.22$_{\pm 2.78}$ & \textbf{44.89$_{\pm 3.43}$} \\
                                & Weighted F1    & \underline{50.43$_{\pm 2.91}$} & 49.20$_{\pm 1.75}$ & 46.36$_{\pm 1.53}$ & 49.36$_{\pm 2.45}$ & \textbf{50.97$_{\pm 2.96}$} \\
    \midrule
    \multirow{3}{*}{Physionet-MI} & Balanced Acc & 60.08$_{\pm 0.16}$ & \underline{60.65$_{\pm 0.98}$} & 60.55$_{\pm 0.96}$ & 58.83$_{\pm 1.35}$ & \textbf{61.15$_{\pm 0.86}$} \\
                                & Cohen's Kappa  & 46.81$_{\pm 0.21}$ & \underline{47.58$_{\pm 1.30}$} & 47.41$_{\pm 1.28}$ & 45.13$_{\pm 1.79}$ & \textbf{48.23$_{\pm 1.15}$} \\
                                & Weighted F1    & 60.17$_{\pm 0.35}$ & \underline{60.66$_{\pm 0.96}$} & 60.47$_{\pm 1.05}$ & 58.83$_{\pm 1.42}$ & \textbf{61.16$_{\pm 0.90}$} \\
    \midrule
    \multirow{3}{*}{Sleep-EDFx} & Balanced Acc & 67.25$_{\pm 3.57}$ & 66.47$_{\pm 2.94}$ & 67.99$_{\pm 3.72}$ & \underline{68.69$_{\pm 3.67}$} & \textbf{70.24$_{\pm 4.63}$} \\
                                & Cohen's Kappa  & 63.94$_{\pm 6.47}$ & 66.81$_{\pm 6.05}$ & \underline{68.13$_{\pm 6.07}$} & 65.33$_{\pm 5.77}$ & \textbf{68.91$_{\pm 6.47}$} \\
                                & Weighted F1    & 74.02$_{\pm 5.06}$ & 75.15$_{\pm 4.63}$ & \underline{76.34$_{\pm 4.35}$} & 75.30$_{\pm 3.99}$ & \textbf{77.31$_{\pm 5.00}$} \\
    \midrule
    \multirow{3}{*}{SHU-MI} & Balanced Acc & \underline{62.04$_{\pm 3.78}$} & 62.01$_{\pm 3.10}$ & 61.74$_{\pm 3.75}$ & 61.08$_{\pm 3.60}$ & \textbf{62.43$_{\pm 3.18}$} \\
                                & AUC-PR  & \textbf{68.84$_{\pm 5.06}$} & 68.38$_{\pm 4.24}$ & 68.28$_{\pm 4.43}$ & 67.11$_{\pm 5.38}$ & \underline{68.73$_{\pm 5.03}$} \\
                             & AUROC  & \textbf{68.14$_{\pm 4.56}$} & 67.81$_{\pm 4.09}$ & 67.46$_{\pm 4.29}$ & 66.76$_{\pm 4.97}$ & \textbf{68.14$_{\pm 5.06}$} \\
    \midrule
    \multicolumn{2}{l}{\textbf{Average Rank}} & \underline{3.00} & 3.25 & 3.58 & 4.00 & \textbf{1.08} \\
    \bottomrule
    \end{tabular}%
}
\end{table*}

\textbf{No single fixed patch size offers a universal solution, as the optimal granularity shifts from task to task. }To validate MAGE's architecture, we conducted an ablation study to assess performance when restricting the model to a single timescale. In this setup, only one patch size is active, negating contributions from others. As shown in Table~\ref{tab:single_granularity}, the performance of each timescale varies heavily depending on the tasks. In FACED dataset, the timescale with patch size 250ms only outperforms the scales with patch size 500ms and patch size 1s. Conversely, on the Sleep-EDFx dataset, the patch size of 1s proved superior to the other two. Across almost all datasets, MAGE achieves a fair or better performance compared to any single timescale. This demonstrates that MAGE can effectively aggregate information from all timescales for each specific task.

\textbf{MAGE-router addresses the trade-off between accuracy and efficiency by dynamically selecting the single most relevant granularity. } While MAGE outperforms each single timescale level, it incurs a higher computational cost. To reduce the this, we introduce MAGE-router, an efficient variant designed for diverse downstream tasks. In this variant, after calculating adaptive weights, only the timescale corresponding to the largest weight is activated, while the others are deactivated. While MAGE-router reduces computational overhead, it experiences a performance drop on certain datasets, at times underperforming even the individual single-scale models (Table~\ref{tab:single_granularity}). We hypothesize this is due to the hard-selection mechanism; activating only one scale per sample may disperse gradient updates. Regardless, by automatically routing signals to their optimal scales, MAGE-router eliminates the need for tedious manual hyperparameter tuning. This highly adaptive, resource-optimized approach justifies the performance trade-off, making MAGE-router an ideal solution for low-latency, energy-efficient applications like wearable devices and real-time BCIs.

\textbf{An aggregator is enough for multi-timescale learning. }To validate the Aggregator's effectiveness and efficiency, we benchmark MAGE against two other multi-timescale methods. The first adopts a uniform distribution across all scales. The second applies Medformer's Inter Attention, where the Aggregator is replaced by a Transformer layer inside each encoder block to facilitate cross-scale interaction by attending to different timescales simultaneously. Under identical settings, MAGE performs better than both methods on the FACED, PhysioNet-MI, and Sleep-EDFx datasets (Table.~\ref{tab:medformer}). Conversely, Inter Attention is most effective on the SHU-MI dataset. To further validate these findings, we statistically compared MAGE against these approaches. Our analysis shows that MAGE significantly outperforms both baselines on two metrics for the Sleep-EDFx dataset, while maintaining comparable performance across the remaining datasets. Notably, since our aggregator contains only a few thousand parameters, MAGE maintains an efficiency comparable to the Uniform Weight method. Meanwhile, integrating additional attention modules within each block, as in Inter Attention, increases model size, particularly when scaling up. Ultimately, these results confirm that our adaptive aggregator not only guarantees stable baseline performance but also provides a distinct, statistically significant advantage without the computational overhead of alternative multi-scale strategies.

\begin{table}[t]
    \centering
    \caption{Comparison of MAGE with other multi-timescale methods. MAGE remains competitive with the Inter Attention method, even while utilizing a lightweight aggregator in contrast to a complex attention module. \textbf{Bold} indicates the best result, and \underline{underlined} indicates the second best. The symbol $\downarrow$ indicates that the baseline model is significantly worse than MAGE.}
    
    \label{tab:medformer}
    \resizebox{0.85\linewidth}{!}{%
        \begin{tabular}{llccc}
            \toprule
            Dataset & Metrics & MAGE & Inter Attention & Uniform Weight \\
            \midrule
            \multirow{3}{*}{FACED} & Balanced Acc & \textbf{51.42}$_{{\pm 2.11}}$ & 50.79$_{{\pm 3.04}}$ & \underline{50.85}$_{{\pm 2.48}}$ \\
                                                   & Cohen's Kappa  & \textbf{44.97}$_{{\pm 2.31}}$ & 44.27$_{{\pm 3.33}}$ & \underline{44.43}$_{{\pm 2.73}}$ \\
                                                   & Weighted F1    & \textbf{50.94}$_{{\pm 1.97}}$ & 50.26$_{{\pm 2.93}}$ & \underline{50.52}$_{{\pm 2.38}}$ \\
            \midrule
            \multirow{3}{*}{PhysioNet-MI} & Balanced Acc & \textbf{62.69}$_{{\pm 3.59}}$ & 60.48$_{{\pm 0.87}}$ & \underline{60.90}$_{{\pm 1.02}}$ \\
                                                   & Cohen's Kappa  & \textbf{50.27}$_{{\pm 4.77}}$ & 47.31$_{{\pm 1.17}}$ & \underline{47.87}$_{{\pm 1.35}}$ \\
                                                   & Weighted F1    & \textbf{62.71}$_{{\pm 3.54}}$ & 60.68$_{{\pm 0.87}}$ & \underline{61.09}$_{{\pm 1.05}}$ \\
            \midrule
            \multirow{3}{*}{Sleep-EDFx} & Balanced Acc & \underline{70.08}$_{{\pm 4.02}}$ & \textbf{70.14}$_{{\pm 3.33}}$ & 69.72$_{{\pm 3.62}}$ \\
                                                   & Cohen's Kappa  & \textbf{68.89}$_{{\pm 5.56}}$ & \underline{66.10}$_{{\pm 6.14}}$$\downarrow$ & 65.56$_{{\pm 6.51}}$$\downarrow$ \\
                                                   & Weighted F1    & \textbf{77.33}$_{{\pm 4.47}}$ & 75.37$_{{\pm 4.49}}$$\downarrow$ & \underline{75.42}$_{{\pm 4.50}}$$\downarrow$ \\
            \midrule
            \multirow{3}{*}{SHU-MI} & Balanced Acc & 62.22$_{{\pm 3.49}}$ & \textbf{62.95}$_{{\pm 3.14}}$ & \underline{62.77}$_{{\pm 2.92}}$  \\
                                                   & AUC-PR  & 68.63$_{{\pm 4.77}}$ & \underline{69.59}$_{{\pm 4.34}}$ & \textbf{69.60}$_{{\pm 4.24}}$ \\
                                                   & AUROC   & 67.91$_{{\pm 4.73}}$ & \textbf{69.01}$_{{\pm 4.15}}$ & \underline{68.95}$_{{\pm 4.17}}$ \\
            \bottomrule
        \end{tabular}%
    }
\end{table}
\textbf{Limitations. }Despite these benefits, there are a few limitations that require attention. First, while MAGE demonstrates strong generalizability, it remains dependent on a pre-defined set of patch sizes as a hyperparameter that requires manual optimization. Our work, however, demonstrates that incorporating a diverse range of patch scales is precisely what drives performance gains, representing a significant stride toward a truly generalizable EEG foundation model. Second, the attention mechanism costs quadratically with sequence length, leading to high computational costs when processing multi-timescale inputs simultaneously. Third, as shown in Figure~\ref{fig:scale_importance}, while both SHU-MI and PhysioNet-MI involve motor imagery, their scale preferences differ. We acknowledge that this difference cannot be fully explained at present. Despite sharing a motor imagery paradigm, the two datasets differ in channel configuration, task labels, cue design, and session structure, which are factors that may affect the temporal characteristics of the learned EEG patterns. One possible reason is that longer windows capture more stable mu/beta modulation, whereas shorter windows capture transient or cue-related activity, and the specific target classes of each dataset dictate which of these temporal features are most critical for classification. Therefore, the learned weights should be viewed as dataset-specific preferences rather than universal properties of motor imagery.

\textbf{Toward generalizable multi-timescale learning. } Lastly, we revisit our core research question: our findings demonstrate that single-scale tokenization is insufficient for building generalized EEG foundation models. By decomposing signals to learn features across distinct timescales and dynamically weighting them, we effectively reduce the architectural complexity required for feature extraction. Consequently, this approach ensures robust generalization across diverse domains while requiring a remarkably small number of parameters. \looseness -1

\section{Conclusion}
In this work, we demonstrate the effectiveness of a multi-timescale architecture in building generalizable EEG foundation models. Our proposed \textbf{MAGE} effectively captures multi-timescale information by segmenting input signals by varying patch sizes. Crucially, MAGE employs a dynamic weighting mechanism, allowing it to adaptively prioritize different timescales based on the specific characteristics of the input signal. Experimental results show that our model achieves superior performance across different datasets spanning different frequency bands.

We believe that multi-timescale architectures for EEG signals remain a promising yet underexplored direction. Currently, while MAGE demonstrates strong generalizability, its reliance on a manually optimized, pre-defined list of patch sizes remains a limitation. Our future work will primarily address this by seeking ways to dynamically generate or select patch scales, removing the need for a fixed list. Additionally, we plan to reduce the computational cost of our attention mechanisms. By doing so, the model will be able to efficiently process a broader, continuous range of temporal information, paving the way for a more robust EEG foundation model. Furthermore, we aim to apply MAGE in a multi-modal context, where EEG can be aligned with other sources of signal like ECG, EMG, EOG~\cite{jiang2025towards}, speech~\cite{fan2025m3anet}, or fMRI~\cite{li2024neurobolt}.

\section*{Acknowledgement}
Part of this work was funded by Hi!PARIS Center on Data Analytics and Artificial Intelligence, by the European Union’s HORIZON Research and Innovation Programme under grant agreement No 101120657, project ENFIELD (European Lighthouse to Manifest Trustworthy and Green AI) and by French National Research Agency (ANR-22-PEFT-0003 and ANR-22-PEFT-0007) as part of France 2030, the NF-NAI project and NF-FITNESS project.\looseness -1
\bibliography{references}
\bibliographystyle{unsrtnat}
\newpage
\appendix

\section*{Appendix Table of Contents}
\startcontents[appendices]
\printcontents[appendices]{l}{1}{\setcounter{tocdepth}{2}}
\vspace{1cm}

\section{Notations}

\begin{longtable}{l l p{10cm}}
    \caption{Table of notations.} \label{tab:notation} \\
    
    \toprule
    \textbf{Symbol} & \textbf{Space} & \textbf{Meaning} \\
    \midrule
    \endfirsthead 

    \multicolumn{3}{c}{{ \tablename\ \thetable{} -- Continued from previous page}} \\
    \addlinespace
    \toprule
    \textbf{Symbol} & \textbf{Space} & \textbf{Meaning} \\
    \midrule
    \endhead 

    \midrule
    \multicolumn{3}{r}{{Continued on next page...}} \\
    \endfoot 

    \bottomrule
    \endlastfoot 

    $C$ & $\mathbb{N}$ & Number of channels \\
    $T$ & $\mathbb{N}$ & Number of timestamps \\
    $G$ & $\mathbb{N}$ & Number of patch sizes \\
    $p_g$ & $\mathbb{N}$ & Patch size at index $g$ \\
    $n_g$ & $\mathbb{N}$ & Number of segments at index $g$ \\ 
    $D$ & $\mathbb{N}$ & Feature dimension \\ 
    $N_{\text{class}}$ & $\mathbb{N}$ & Number of target classes for downstream task \\
    \addlinespace
    
    $\mathbf{E}$ & $\mathbb{R}^{C\times T}$ & Input signal \\
    $\hat{\mathbf{E}}$ & $\mathbb{R}^{C\times T}$ & Reconstructed signal \\
    $\mathbf{E}_{p_g}$ & $\mathbb{R}^{C\times n_g \times p_g}$ & Signal after patching with patch size $p_g$ \\
    $\mathbf{e}_{p_g}$ & $\mathbb{R}^{p_g}$ & One patch at patch size $p_g$ \\ 
    $\mathbf{x}_{p_g}$ & $\mathbb{R}^{D}$ & One token at patch size $p_g$ \\ 
    \addlinespace
    
    $\mathbf{a}_{\text{temp}}$ & $\mathbb{R}^{D_h}$ & Aggregator temporal vector \\
    $\mathbf{a}_{\text{spec}}$ & $\mathbb{R}^{D_h}$ & Aggregator spectral vector \\
    $\mathbf{a}$ & $\mathbb{R}^{2 D_h}$ & Aggregator features vector \\
    $\mathbf{\alpha}$ & $\mathbb{R}^{G}$ & Deterministic output of aggregator \\
    $\tilde{\mathbf{\alpha}}$ & $\mathbb{R}^{G}$ & Stochastic output of aggregator \\
    $\mathbf{\epsilon}$ & $\mathbb{R}^{G}$ & Gaussian noise vector \\
    $\sigma$ & $\mathbb{R}$ & Standard deviation of Gaussian noise \\
    \addlinespace
    
    $\mathbf{X}^{(g)}$ & $\mathbb{R}^{C\times n_g \times D}$ & Tokens at timescale $g$ \\
    $\mathbf{M}^{(g)}$ & $\mathbb{R}^{C\times n_g \times D}$ & Masks at timescale $g$ \\
    \addlinespace
    
    $\mathbf{X}_c$ & $\mathbb{R}^{n_g \times D}$ & Token at channel $c$ \\
    $\mathbf{W}^T_Q$ & $\mathbb{R}^{D\times D}$ & Query projection for temporal attention \\
    $\mathbf{W}^T_K$ & $\mathbb{R}^{D\times D}$ & Key projection for temporal attention \\
    $\mathbf{W}^T_V$ & $\mathbb{R}^{D\times D}$ & Value projection for temporal attention \\
    $\mathbf{W}^T_O$ & $\mathbb{R}^{D\times D}$ & Output projection for temporal attention \\
    $\mathbf{Q}_c$ & $\mathbb{R}^{n_g\times D}$ & Query matrix for temporal attention \\
    $\mathbf{K}_c$ & $\mathbb{R}^{n_g\times D}$ & Key matrix for temporal attention \\
    $\mathbf{V}_c$ & $\mathbb{R}^{n_g\times D}$ & Value matrix for temporal attention \\
    $\mathbf{F}_c$ & $\mathbb{R}^{n_g\times D}$ & Output for temporal attention at channel $c$ \\
    $\mathbf{F}^{(g)}$ & $\mathbb{R}^{C\times n_g\times D}$ & Output for temporal attention \\
    \addlinespace
    
    $\mathbf{F}_t$ & $\mathbb{R}^{C \times D}$ & Token at time $t$ \\
    $\mathbf{W}^S_Q$ & $\mathbb{R}^{D\times D}$ & Query projection for spatial attention \\
    $\mathbf{W}^S_K$ & $\mathbb{R}^{D\times D}$ & Key projection for spatial attention \\
    $\mathbf{W}^S_V$ & $\mathbb{R}^{D\times D}$ & Value projection for spatial attention \\
    $\mathbf{W}^S_O$ & $\mathbb{R}^{D\times D}$ & Output projection for spatial attention \\
    $\mathbf{Q}'_t$ & $\mathbb{R}^{C\times D}$ & Query matrix for spatial attention \\
    $\mathbf{K}'_t$ & $\mathbb{R}^{C\times D}$ & Key matrix for spatial attention \\
    $\mathbf{V}'_t$ & $\mathbb{R}^{C\times D}$ & Value matrix for spatial attention \\
    $\mathbf{U}_t$ & $\mathbb{R}^{C\times D}$ & Output for spatial attention at time $t$ \\
    $\mathbf{U}^{(g)}$ & $\mathbb{R}^{C\times n_g\times D}$ & Output for spatial attention \\
    \addlinespace
    
    $\mathbf{Z}^{(g)}_S$ & $\mathbb{R}^{C\times n_g\times D}$ & Student latent representation at timescale $g$ \\
    $\mathbf{H}^{(g)}_T$ & $\mathbb{R}^{C\times n_g\times D}$ & Teacher target representation at timescale $g$ \\
    $\hat{\mathbf{H}}^{(g)}_S$ & $\mathbb{R}^{C\times n_g\times D}$ & Student predicted representation at timescale $g$ \\
    $\hat{\mathbf{y}}$ & $\mathbb{R}^{N_{\text{class}}}$ & Final task prediction vector \\
    \addlinespace
    
    $\mathcal{L}_{\text{align}}$ & $\mathbb{R}$ & Alignment loss \\
    $\mathcal{L}_{\text{recon}}$ & $\mathbb{R}$ & Signal reconstruction loss \\
    $\mathcal{L}$ & $\mathbb{R}$ & Total objective function \\
    $\lambda_{\text{recon}}$ & $\mathbb{R}$ & Weight hyperparameter for reconstruction loss \\
    $\lambda_{\text{align}}$ & $\mathbb{R}$ & Weight hyperparameter for alignment loss \\
    \addlinespace
    
    $\mathcal{P}$ & Set & Set of patch sizes \\
    $\mathcal{X}_\mathcal{P}$ & Set & Set of input tokens \\
    $\tilde{\mathcal{X}}_\mathcal{P}$ & Set & Set of masked input tokens \\
    $\mathcal{Z}_S$ & Set & Set of student latent representations \\
    $\mathcal{H}_T$ & Set & Set of teacher target representations \\
    $\hat{\mathcal{H}}_S$ & Set & Set of student predicted representations \\

\end{longtable}

\section{Additional Pretraining setup}
\label{appendix: pretrain}
\textbf{Pretraining dataset. }MAGE is pretrained on a subset of 500 hours of TUEG. TUEG is recorded using a standard 10-20 system, with most of the data sampled at 256 Hz. To avoid data leaking, the TUAB dataset is rigorously excluded from the pretraining subset. The TUEG dataset suffers from significant data contamination, including substantial unannotated noise, artifacts, and channel mismatch. To address this, we utilize an efficient preprocessing strategy designed to filter noise from the pre-training data.

\textbf{Preprocessing. }Firstly, the beginning and ending of each TUEG recording are contaminated. Therefore, in each recording, the first and last one minute are removed to avoid low-quality data. Secondly, to exclude degraded channels while preserving crucial information, we restricted our dataset to the 19 most standard electrode positions: Fp1, Fp2, F7, F3, Fz, F4, F8, T3, C3, Cz, C4, T4, T5, P3, Pz, P4, T6, O1, and O2. After that, we apply a band-pass filter (0.3 Hz - 75 Hz) to remove noise, along with a notch filter at 60 Hz to discard the power line noise. All recordings are then resampled to 200 Hz, and cut into 30-second windows to create uniformly long-term samples. Raw EEG signals are frequently contaminated by non-brain physiological artifacts, such as ECG, EOG, and EMG~\cite{yin2023gan}. These artifacts can introduce high-amplitude pulses that mimic abnormal brain activity, potentially misleading the model. We filtered out any samples with voltage amplitudes exceeding 100 $\mu$V because such high-magnitude peaks deviate from standard physiological brain activity and typically represent non-cerebral artifacts. Finally, we normalized each sample by z-score normalization per channel to standardize the data.

\begin{table*}[t]
\centering
\caption{Hyperparameters for MAGE pre-training on a single NVIDIA A100 40GB.}
\label{tab:hparams}
\begin{tabularx}{\textwidth}{llX}
\toprule
\textbf{Component} & \textbf{Hyperparameters} & \textbf{Settings} \\
\midrule
EEG sample & Channels & 19  \\
           & Time points & 6000 (30s @ 200Hz) \\
           & Patch sizes & 50 100 200 (250ms 500ms 1s) \\
           & Mask ratio & 0.7 \\
           & Mask token & Constant \\
\midrule
Encoder & Layers (depth) & 6 \\
           & Hidden dimension & 64  \\
           & Attention heads & 4 \\
           & MLP ratio & 4.0 \\
           & $qkv$ bias & True \\
           & Init std & 0.02 \\
           & Norm $\epsilon$ & $1\times10^{-6}$ \\
           & Attention pattern & Multi-timescale Attention \\
\midrule
Predictor & Depth & 3 \\
          & Attention & Multi-timescale Attention \\
\midrule
Aggregator & Architecture & Dual-Stream (Temporal + Frequency) \\
          & Temporal Dim & 32 \\
          & Frequency Dim & 16 \\
          & Temporal Extractor & $2\times \text{Conv1D} (k=\{7, 5\}, s=4)$, BatchNorm, GELU, AdaptiveAvgPool \\
          & Magnitude Extractor & 2$\times$Conv1D ($k=\{5, 3\}, s=2$), BN, GELU, AdaptiveAvgPool \\
          & Phase Extractor & 1$\times$Conv1D ($k=5, s=2$), BN, GELU, AdaptiveAvgPool \\
          & Fusion & Cross-domain Linear Projection \\
\midrule
Reconstructor & Head & Linear (hidden dimension, patch dimension) \\
\midrule
Momentum Encoder & EMA momentum & Cosine ramp $0.996 \to 0.9995$ \\
\midrule
Pre-training & Epochs & 50 \\
             & Batch size & 64 (global) \\
             & Optimizer & AdamW \\
             & Learning rate & $1\times10^{-4}$ \\
             & LR schedule & Warmup + Cosine decay \\
             & Warmup steps & $\max(500, 0.1T)$ \\
             & Weight decay & $5\times10^{-2}$ (cosine schedule) \\
             & Adam $\beta$ & (0.9, 0.999) \\
             & Adam $\epsilon$ & $1\times10^{-8}$ \\
             & Gradient clipping & 1.0 \\
             & Loss weights & $\lambda_{\text{rec}}=5.0$, $\lambda_{\text{align}}=1.0$ \\
             & Num worker & 32 \\
\bottomrule
\end{tabularx}
\end{table*}

\textbf{Pretraining settings. }Our experiments are implemented with Pytorch 2.1.2 and CUDA 11.8. During each training iteration, we select a set of patch sizes: 50, 100, and 200 (representing 250ms, 500ms, and 1s) to encourage the encoder to learn robust representations across multiple timescales. $70\%$ of patches are randomly masked during the pretraining phase. We adopt the AdamW optimizer with a learning rate $1e-4$, batch size 64, and 50 training epochs. Additional hyperparameters for pretraining MAGE are described in Tab.~\ref{tab:hparams}. The pretraining process for 500 hours of EEG data in MAGE requires approximately 8 hours to complete.
\section{Additional downstream setup}

\subsection{Downstream pipeline}
\label{appendix: downstream}
\begin{figure}
    \centering
    \includegraphics[width=0.49\linewidth]{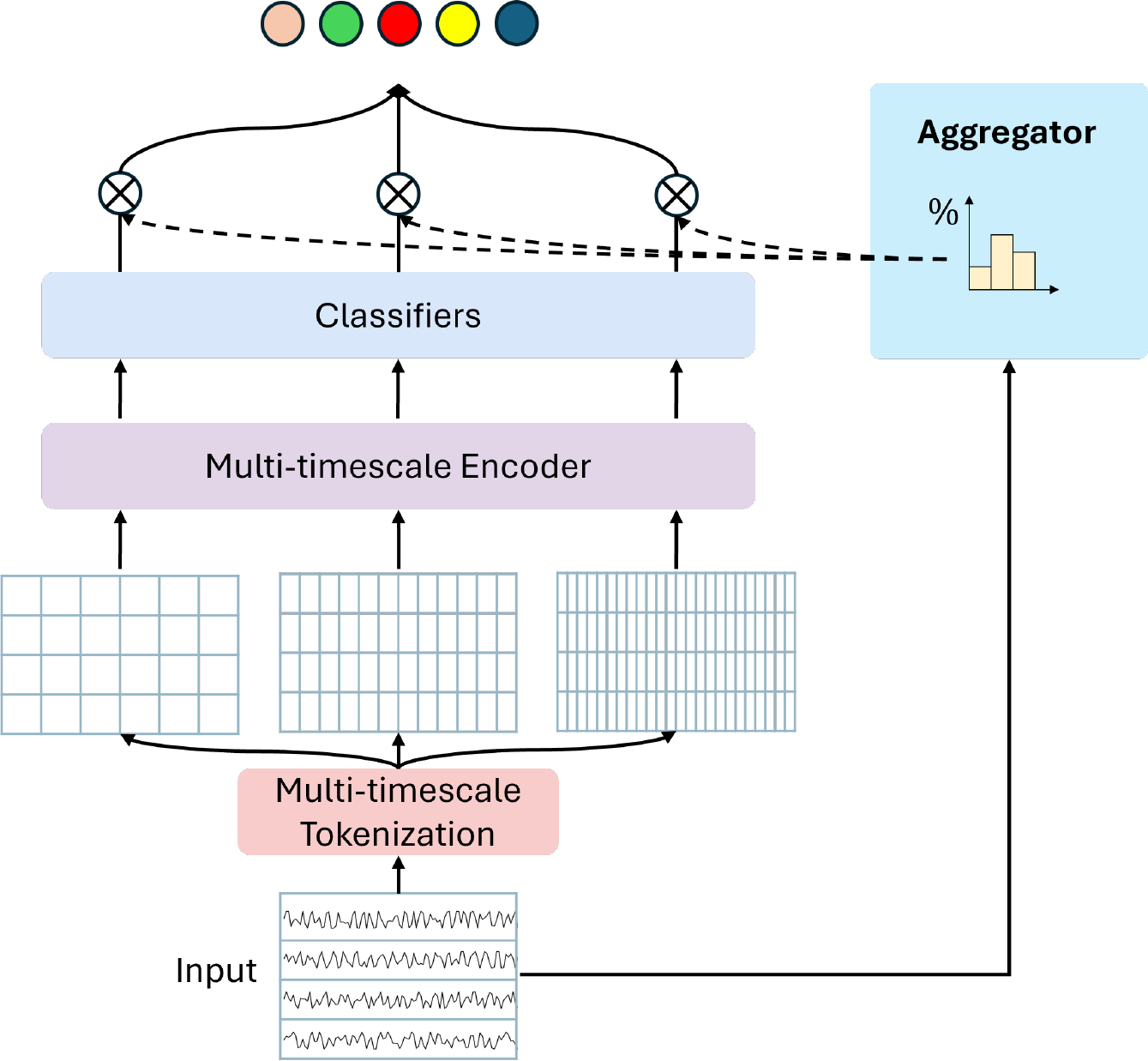}
    \caption{Downstream pipeline of MAGE. Aggregator is adopted to dynamically weight each time scale before adding them together for final prediction.}
    \label{fig:downstream}
\end{figure}

Figure~\ref{fig:downstream} illustrates the adaptation of MAGE for downstream tasks. When downstreaming, only the encoder is retained for evaluation, while the predictor and decoder are discarded. The input signal first passes through a multi-timescale module, which tokenizes the signal into patches of varying sizes. These tokens are processed by the Multi-timescale Encoder to capture information across both timescales and spatial dimensions. An aggregator then determines an importance score for each scale based on temporal and spectral analysis. Similar to the pre-training phase, stochastic noise is introduced during downstream adaptation to prevent the model from over-relying on any single timescale. 

Following the evaluation protocol established by EEGPT, a convolutional layer is employed to project the variable number of input channels from the downstream dataset into a fixed 19-channel space. We then freeze the parameters of the pre-trained encoder and train the tokenization module and the classifiers. This strategy mitigates the risk of overfitting on smaller downstream datasets and enables a direct evaluation of the quality of the representations learned during pre-training. For each timescale $g$, the classification head $\mathrm{Classifier}^{(g)}$ consists of a projection layer followed by a linear classifier. The tokens after the encoder are first flattened across their channel and embedding dimensions. They are subsequently projected to a lower dimension with dropout and max-norm constraints, then concatenated across time steps before passing through the final linear layer. Crucially, this same evaluation head is applied across all baseline models to ensure a fair comparison.

The final task prediction $\hat{\mathbf{y}}$ of MAGE is computed by aggregating these representations via a Classifier:

\begin{align}
    \hat{\mathbf{y}} = \sum_{g=1}^G \tilde{\alpha}_g \cdot \mathrm{Classifier}^{(g)}\left( \mathbf{Z}^{(g)}_S \right),
\end{align}

where $\text{Classifier}^{(g)}$ denotes the task-specific classification head for timescale $g$, and $\tilde{\boldsymbol{\alpha}}$ represents the stochastic importance vector derived from the aggregator. Notably, this stochasticity is disabled during evaluation to ensure
deterministic predictions. 

\subsection{Datasets}
\label{appendix: dataset}
When evaluating the EEG model, performance is strongly affected by the partition of the subject, which makes the result remarkably different between different selections. Relying on a fixed division of training, validation and testing sets prevents the foundation models from revealing their actual potential. Therefore, we adopt a subject-wise cross-validation evaluation scheme, in which, for each dataset, we evaluate each dataset $N$ folds with different partitions of train, val, and test sets. The reported performance is the average across all folds. Detailed information about the train-validation-test split is illustrated in each dataset below.

\begin{table*}
\centering
\caption{Overview of evaluation EEG datasets.}
\label{tab:downstream-datasets}
\begin{tabular}{lcccc}
\toprule
\textbf{Dataset} & \textbf{Rate} & \textbf{Number of Channels} & \textbf{Duration} & \textbf{Labels} \\ \midrule
    BCIC-IV-2A       & 250 Hz   & 22     & 4s    & 4-class  \\
    PhysioNet-MI & 160 Hz   & 64     & 4s    &  4-class\\ 
    FACED & 250 Hz & 32 & 10s & 9-class \\ 
    Sleep-EDFx     & 100 Hz   & 2      & 30s  & 5-class \\ 
    TUAB         & 250 Hz & 16      & 10s & 2-class \\ 
    KaggleERN    & 200 Hz   & 56      &2s     & 2-class \\
    SHU-MI & 250 Hz & 32 & 4s & 2-class \\
\bottomrule
\end{tabular}
\end{table*}

\subsubsection{BCIC-IV-2a}
\textbf{License: }The dataset is sourced from the BNCI Horizon 2020 repository and is utilized under the Creative Commons Attribution No Derivatives (CC BY-ND 4.0) license.

\textbf{Dataset Overview and Preprocessing. }BCIC-IV-2a consists of recordings from nine subjects, each subject performs four different motor imagery tasks: left hand, right hand, feet, and tongue. Each subject conducted two different sessions on different days. Each session contains 288 trials. The data is preprocessed with a 0-38Hz band-pass filter and resampled to 200Hz. Finally, for each 4-second trial, the z-score normalization per channel is applied to standardize the data. 

\textbf{Evaluation. }We utilize a leave-one-subject-out cross-validation method. We evaluate each model nine times, with each subject being a test subject each time. 

\subsubsection{PhysioNet-MI}
\textbf{License: }The dataset is made available by PhysioNet under the Open Data Commons Attribution License (ODC-By 1.0).

\textbf{Dataset Overview and Preprocessing. }PhysioNet-MI contains motor imagery recordings from 109 subjects. The task includes four different classes: left fist, right fist, both fists, and both feet. The data is preprocessed with a 0.3Hz cutoff frequency high-pass filter, resampled to 200Hz, and segmented to 4-second samples. We then apply z-score normalization per channel to each sample to standardize the data.

\textbf{Evaluation. }We consider the evaluation set of PhysioNet-MI as the test set. For its train set, we split it into five folds with different evaluation set each fold. The models will then be evaluated five times with different train and evaluation sets. 

\subsubsection{FACED}
\textbf{License: }The dataset is published in Scientific Data and is used under the Creative Commons Attribution 4.0 International (CC BY 4.0) license.

\textbf{Dataset Overview and Preprocessing. }The FACED dataset comprises 32-channel EEG recordings from 123 subjects watching 28 video clips. The stimuli elicit nine fine-grained emotion classes, including amusement, fear, and neutral. We resample each trial to 6,000 points and segment it into three non-overlapping windows, resulting in samples of 2,000 data points ($32 \times 2000$). Finally, we apply z-score normalization per channel to standardize the data.

\textbf{Evaluation. }We perform subject-independent 4-fold cross-validation. In each fold, one partition is held out for testing. The remaining subjects are further split into training and validation sets to monitor performance during training.

\subsubsection{Sleep-EDFx}
\textbf{License: }The dataset is obtained from the PhysioNet repository and is used under the Open Data Commons Attribution License (ODC-By) v1.0.

\textbf{Dataset Overview and Preprocessing. }The Sleep-EDFx dataset comprises sleep recordings from 78 healthy subjects, categorized into five distinct sleep stages: Wake (W), N1, N2, N3, and REM. We apply a 30 Hz low-pass filter, resample the data to 200 Hz, and segment it into 30-second samples. After that, to standardize the signal, the z-score normalization per channel is applied.

\textbf{Evaluation. }We employ a subject-wise 5-fold cross-validation strategy. The subjects are equally divided into five folds. In each test, one fold serves as the test set, while the remaining four folds are used for training and validation. These training subjects are further split with a 9:1 ratio for training and validation.

\subsubsection{TUAB}
\textbf{License: }The dataset is provided by the Neural Engineering Data Consortium (NEDC) and is used under the NEDC Community Resource Free Academic Use License.

\textbf{Dataset Overview and Preprocessing. }The TUAB dataset contains 409,455 10-second EEG samples, labeled for binary classification as either normal or abnormal. We apply a 0.1–75 Hz band-pass filter and a 50 Hz notch filter to remove line noise. The data is resampled to 200 Hz and applied a z-score normalization per channel to standardize the signal.

\textbf{Evaluation. }We utilize the official evaluation set provided by TUAB as our held-out test set. For training and validation, we adopt a 4-fold cross-validation protocol on the remaining subjects. In each run, one fold serves as the validation set while the other three form the training set, resulting in an overall train-validation-test ratio of approximately 6:2:2.

\subsubsection{KaggleERN}
\textbf{License: }The dataset was originally released for the BCI Challenge @ NER 2015 and is used in accordance with the Kaggle Competition Rules for academic and non-commercial research.

\textbf{Dataset Overview and Preprocessing. }KaggleERN captures error-related potentials (ErrP) from subjects using a P300 speller. This is a binary classification task indicating whether the system's feedback matched the user's intent (Correct vs. Erroneous). We resample the signals to 200 Hz and extract 2-second windows per sample. In each sample, we apply the z-score normalization to standardize the signal.

\textbf{Evaluation. }We utilize the Kaggle evaluation set as the test set. For the training data, we employ a subject-wise 5-fold cross-validation method, where one fold serves as a validation set and the rest as a training set.

\subsubsection{SHU-MI}
\textbf{License: }The dataset is used under the Creative Commons Attribution 4.0 International (CC BY 4.0) license.

\textbf{Dataset Overview and Preprocessing. }The SHU-MI dataset contains EEG recordings from 55 subjects performing two distinct motor imagery tasks: left-hand and right-hand movement. We resample each trial to 800 data points (corresponding to a 4-second window at 200 Hz). We then apply z-score normalization per channel to standardize the signal distribution.

\textbf{Evaluation. }We employ a subject-independent 5-fold cross-validation strategy. In each iteration, $20\%$ of the subjects are reserved for testing. The remaining $80\%$ of subjects are further partitioned into a training and validation set with a ratio of 9:1.

\subsection{Baseline model}
\label{appendix: model}

\begin{table}[H]
    \centering
    \caption{The sampling rate and number of parameters of each model.}
    \label{tab:baseline_model}
        \begin{tabular}{lccc}
            \toprule
            \textbf{Models} & \makecell{\textbf{Training} \\ \textbf{Paradigm}} & \makecell{\textbf{Sampling} \\ \textbf{rate (Hz)}} & \makecell{\textbf{Params.} \\ \textbf{(M)}} \\
            \midrule
            Medformer & supervised & 200 & 1.4 \\
            \midrule
            LaBraM & \makecell{self-\\supervised} & 200 & 5.8 \\
            BIOT & \makecell{self-\\supervised} & 200 & 3.2 \\
            EEGPT & \makecell{self-\\supervised} & 256 & 25.0 \\
            CBraMod & \makecell{self-\\supervised} & 200 & 4.0 \\
            CSBrain & \makecell{self-\\supervised} & 200 & 8.9 \\
            \midrule
            MAGE & \makecell{self-\\supervised} & 200 & 0.6 \\
            \bottomrule
        \end{tabular}
\end{table}

\subsubsection{Medformer}
Medformer is a multi-timescale patching transformer model specialized for medical time series data. Medformer learns complex temporal and spatial dependencies by segmenting the input into various patch sizes via cross-channel patching. This multi-scale approach effectively simulates different frequency bands and captures both fine-grained local patterns and coarser global trends. To aggregate this information, the model adopts a two-stage transformer layer: it first applies intra-scale attention to capture features within each scale, and then utilizes inter-scale attention to fuse information across the different patch sizes.

In our experiments, we adopt Medformer with three scales: 250ms, 500ms, 1s, similar to MAGE, and the feature size of the model is 64, resulting in a total of 1.4 M parameters.

\subsubsection{LaBraM}
LaBraM is a large brain foundation model for the EEG signal. It learns EEG generic representations by predicting the discrete neural tokens of masked EEG patches based on a full-attention transformer. LaBraM adopts a novel vector-quantized neural spectrum prediction (VQ-NSP) framework. It first applies short-time Fourier transforms (STFT) to transform the signal into frequency-domain spectrograms and discretizes this signal into neural codes. LaBraM is pretrained on 2500 hours of EEG data with masked reconstruction tasks.

LaBraM has three variants: LaBraM-Base, LaBraM-Large, LaBraM-Huge, with the number of parameters being 5.8M, 46M, and 369M, respectively. Since only the pretrained checkpoint for LaBraM-Base is available, we adopt this variant for our experiments. To ensure a fair comparison with other models, in downstream evaluation, we freeze all encoder layers of LaBraM, while still fine-tuning the patch embedding layer initially to help it adapt to a new model's data domain. Furthermore, all data are resampled to 200 Hz when training with LaBraM to match the pretraining data. 

\subsubsection{BIOT}
BIOT is a pretrained foundation model for EEG signal. It proposes a novel biosignal tokenization module that segments each channel separately into fixed-length tokens and flattens them to form unified "biosignal sentences". BIOT is pre-trained on multiple large-scale datasets with up to 10 million EEG samples in total.

In our experiments, we adopt the BIOT version with 3.2M parameters and the available pretrained checkpoint from 6 datasets with 18 EEG channels. A convolution layer is applied to the input into 18 channels. Similar to other models, we freeze the encoder of BIOT while unfreezing other modules during downstreaming to evaluate the encoded representation of the model. The trained modules include channel convolution, patch embedding, channel tokens, and a classifier head. All data are resampled to 200 Hz to train with BIOT to match the pretraining data. 

\subsubsection{EEGPT}
EEGPT is a pretrained transformer model for universal EEG feature extraction. It proposes a dual self-supervised learning method with learning by both alignment task and reconstruction task during pretraining. While the reconstruction task helps them to learn detailed features, the alignment task support learn rich semantic information. EEGPT is trained on multiple open-source datasets with more than 200 subjects in total.

EEGPT adopts two variants: EEGPT and EEGPT-Tiny, with the number of parameters is 25M and 4.7M, respectively. In our experiments, we adopt the EEGPT variant with available pretrained checkpoint. Similar to other models, we freeze the encoder of EEGPT while unfreezing other modules during downstreaming to evaluate the encoded representation of the model. The trained modules include channel convolution, patch embeddingto , and a classifier head. Notably, while EEGPT is trained on 256Hz data, all other models are trained on 200Hz data. We therefore preprocess to resample all datasets to 256 Hz when evaluating EEGPT.

\subsubsection{CBraMod}
CBraMod is a robust EEG foundation model designed to tackle the structural heterogeneity of brain signals across different datasets. It addresses the full EEG modeling of prior foundation models by proposing a criss-cross attention. In this attention mechanism, CBraMod models spatial and temporal dependencies separately by two parallel attention modules. By doing this, CBraMod can avoid attending information from irrelevant tokens. CBraMod is pretrained by the masked reconstruction task on 9000 hours from the TUEG dataset. 

In our experiments, we adopt CBraMod with 4.0M parameters and an available pretrained checkpoint. Similar to other models, we freeze the encoder of CBraMod while finetuning other modules during downstream evaluation to evaluate the encoded representation of the model. The trained modules include patch embedding and a classifier head. All inputs were resampled to 200 Hz to match CBraMod's pre-training specifications.

\subsubsection{CSBrain}
CSBrain is a foundation model designed to advance general-purpose EEG decoding by capturing the multi-scale characteristics of brain signals. CSBrain introduces a cross-scale spatial-temporal transformer to model the complex spatial correlations between brain regions and dynamic temporal characteristics. CSBrain is pretrained by a masked reconstruction task on 9000 hours of data from the TUEG dataset. 

In our experiments, we utilize the published pretrained checkpoint of CSBrain with 8.9M parameters. We freeze the encoder of CSBrain and fine-tune other modules to evaluate the representation from the encoder. The trained modules include patch embedding and a classifier head. All data were resampled to 200 Hz to align with the pre-training sampling rate of CSBrain.

\subsection{Evaluation metrics}
\label{appendix: metrics}
This section introduces all metrics we employed to evaluate the model's performance.

\textbf{Balanced Accuracy} is defined as the mean of recall for each class in the dataset. Balanced accuracy ensures that the model’s performance is not biased towards the majority class, which is a common issue in imbalanced datasets.

\textbf{Cohen's Kappa} is a statistical metric used to measure the level of agreement between two classifiers. It accounts for the agreement occurring by chance. In the context of the experiments, one classifier is the true label of the sample, and the other is the predicted label. It is used for multiclass classification problems.

\textbf{Weighted F1} score is the average of the F1-scores for all classes, where each class’s score is weighted by its number of true instances. It is used for multiclass classification problems.

\textbf{AUROC} (Area Under the Receiver Operating Characteristic Curve) measures the trade-off between the true positive rate and the false positive rate across all possible classification thresholds. It is used for binary classification problems. 

\textbf{AUC-PR} (Area Under the Precision-Recall Curve) measures the trade-off between precision and recall across different thresholds. It is used for binary classification problems. 

\section{Additional results}
\subsection{Sample-level analysis of dynamic routing}
\label{appendix: sample level analysis}
\begin{figure}[htbp]
\centering

\begin{subfigure}[b]{0.48\textwidth}
    \centering
    \includegraphics[width=\textwidth]{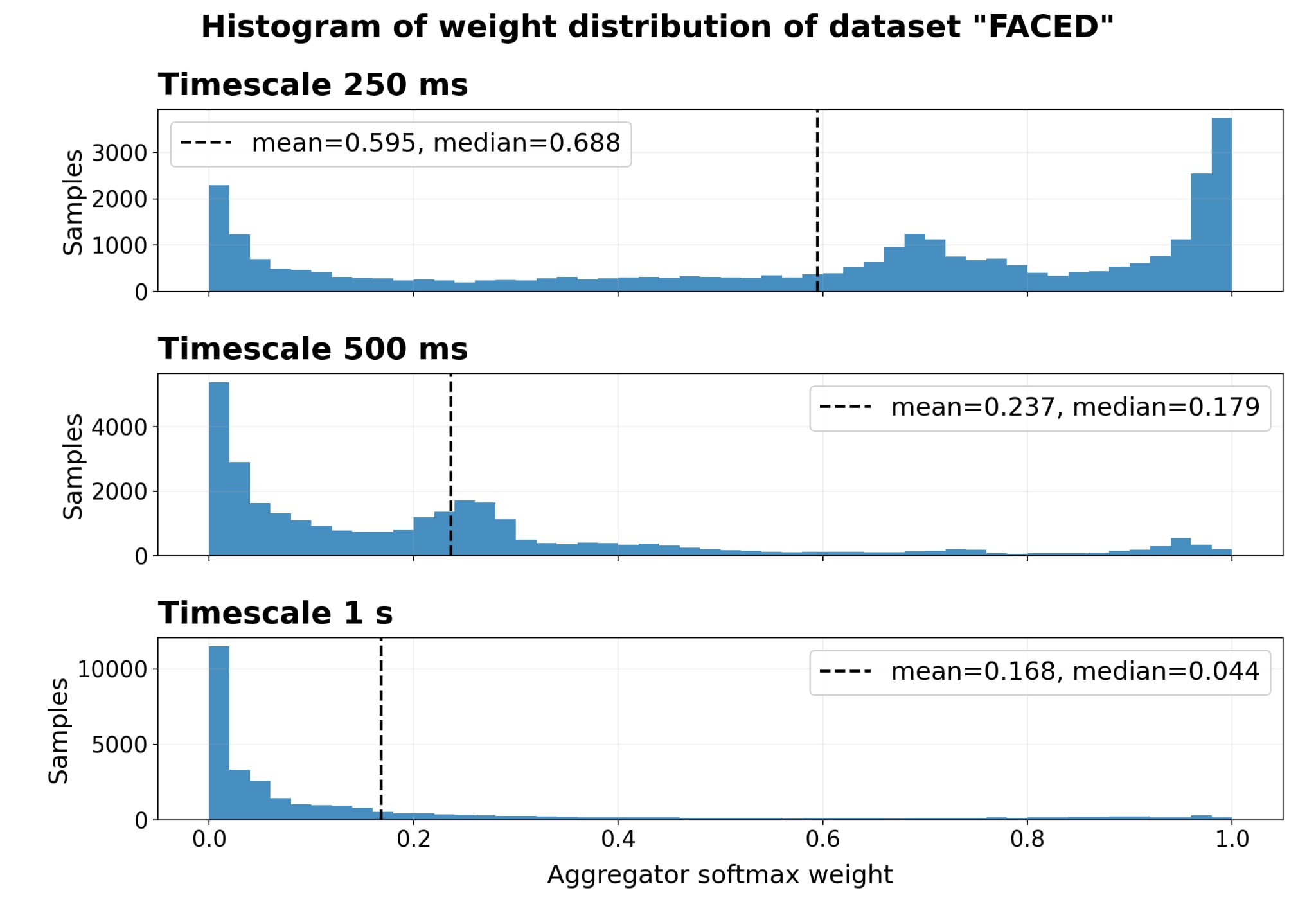}
    \caption{FACED}
    \label{fig:faced}
\end{subfigure}
\hfill
\begin{subfigure}[b]{0.48\textwidth}
    \centering
    \includegraphics[width=\textwidth]{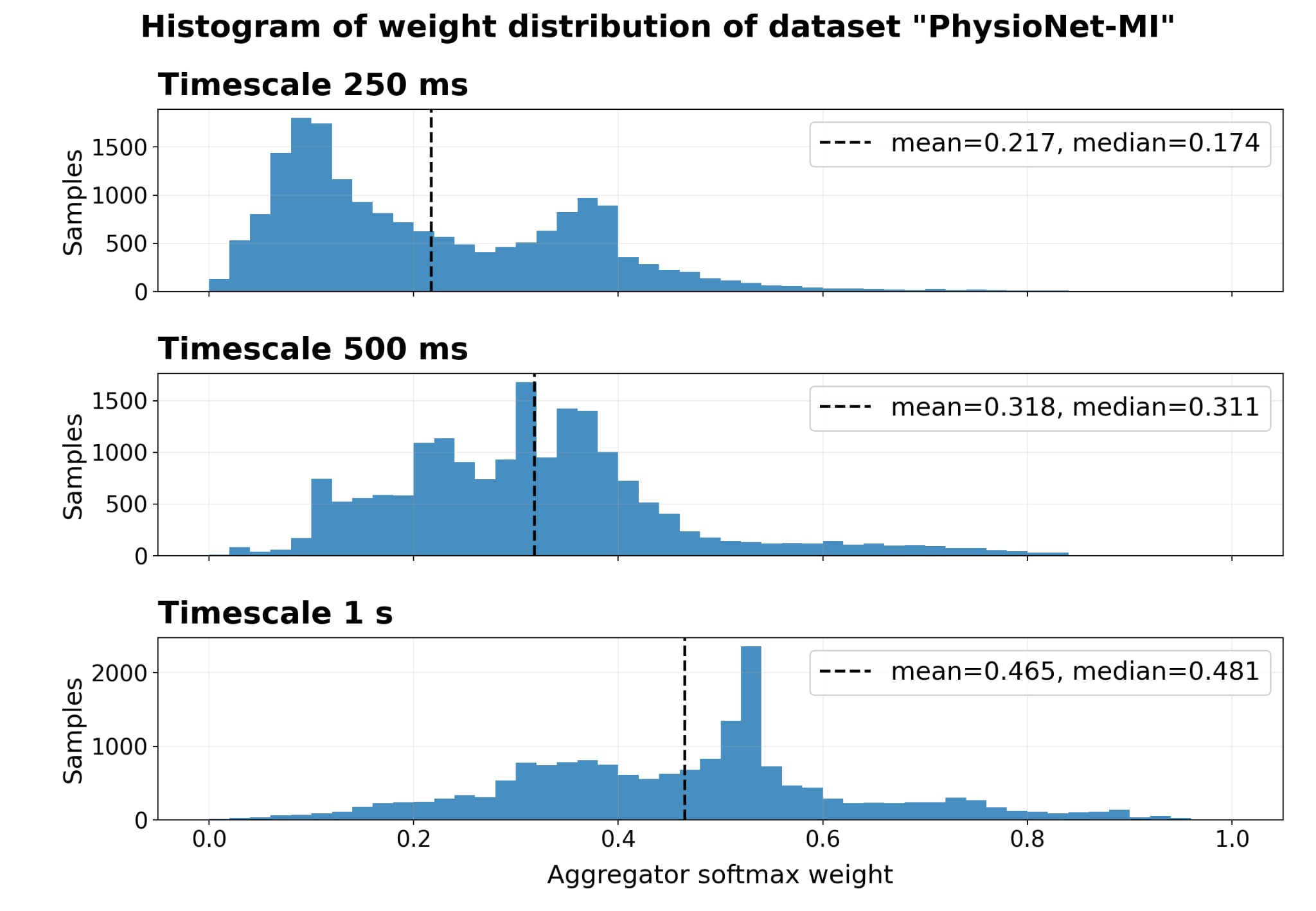}
    \caption{PhysioNet-MI}
    \label{fig:physio}
\end{subfigure}

\begin{subfigure}[b]{0.48\textwidth}
    \centering
    \includegraphics[width=\textwidth]{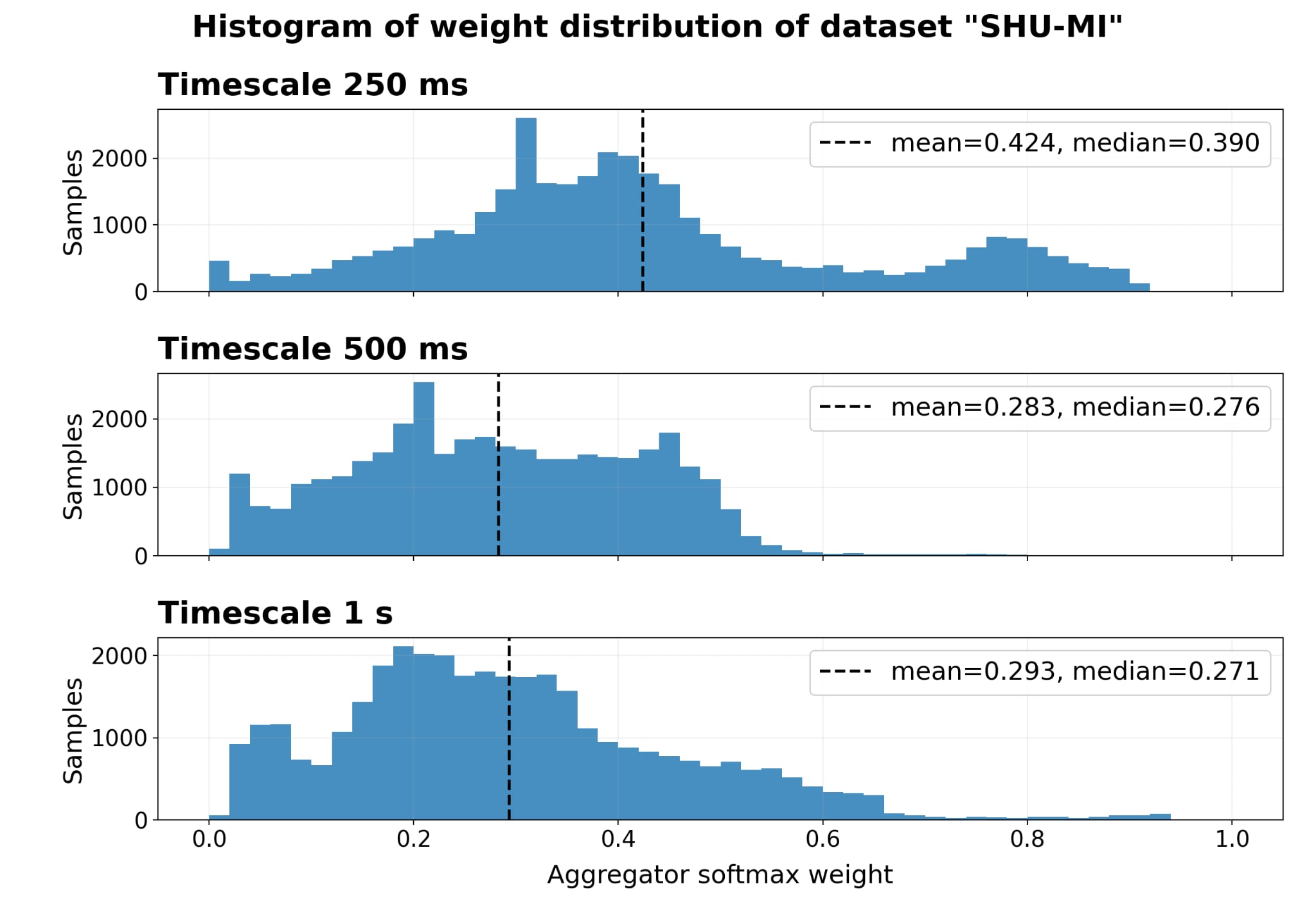}
    \caption{SHU-MI}
    \label{fig:shumi}
\end{subfigure}
\hfill
\begin{subfigure}[b]{0.48\textwidth}
    \centering
    \includegraphics[width=\textwidth]{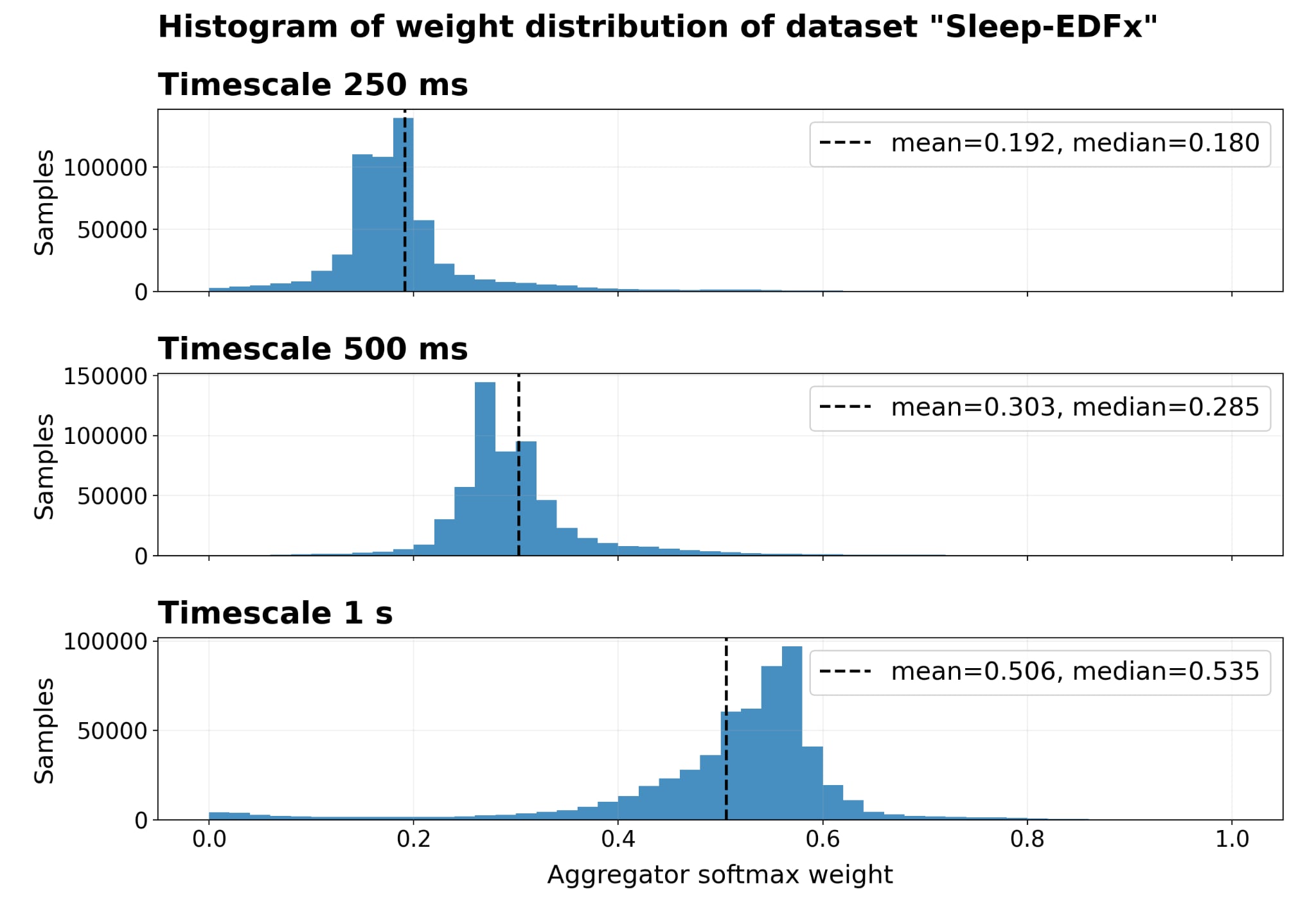}
    \caption{Sleep-EDFx}
    \label{fig:sleepedfx}
\end{subfigure}

\begin{subfigure}[b]{0.48\textwidth}
    \centering
    \includegraphics[width=\textwidth]{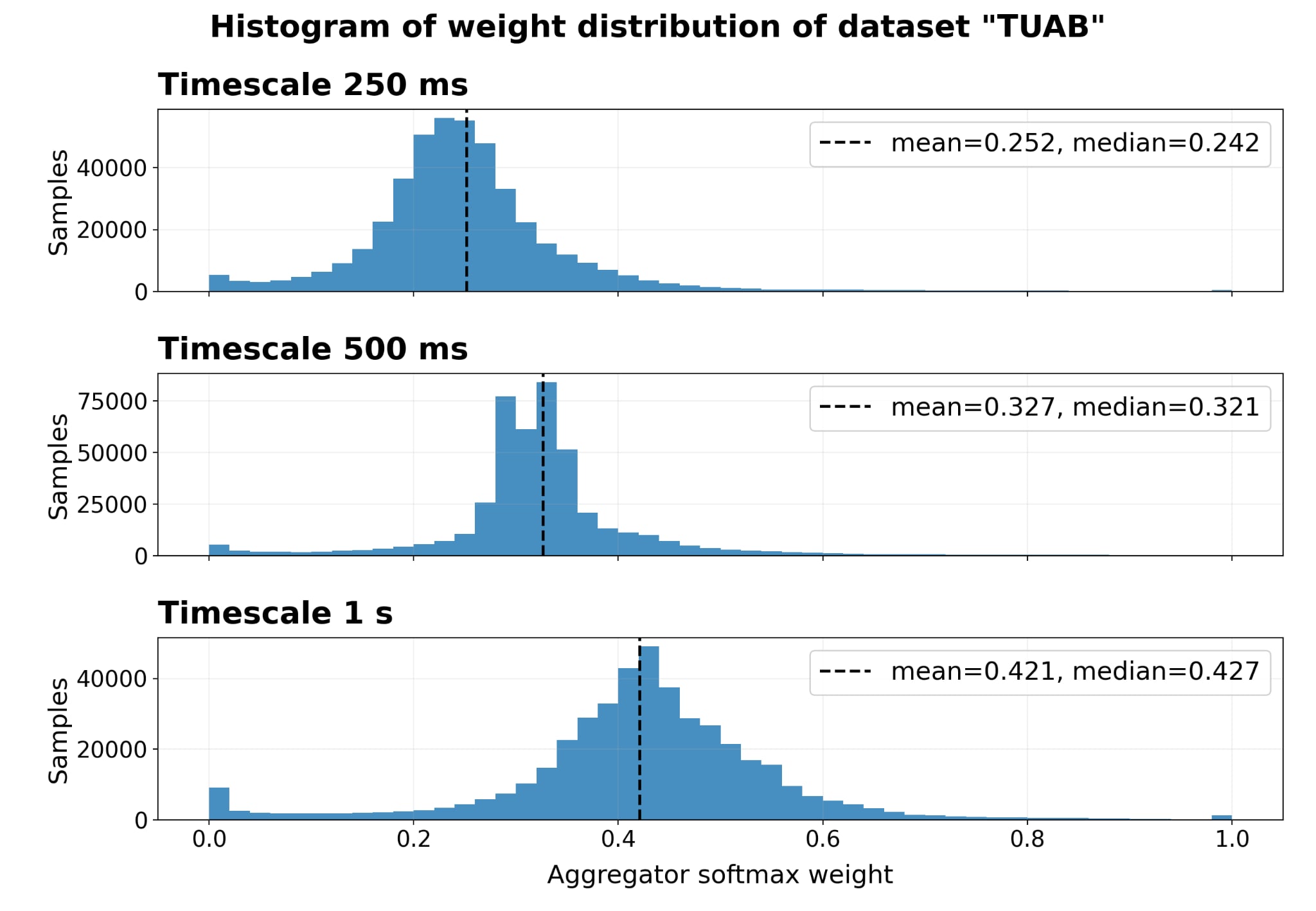}
    \caption{TUAB}
    \label{fig:tuab}
\end{subfigure}
\hfill
\begin{subfigure}[b]{0.48\textwidth}
    \centering
    \includegraphics[width=\textwidth]{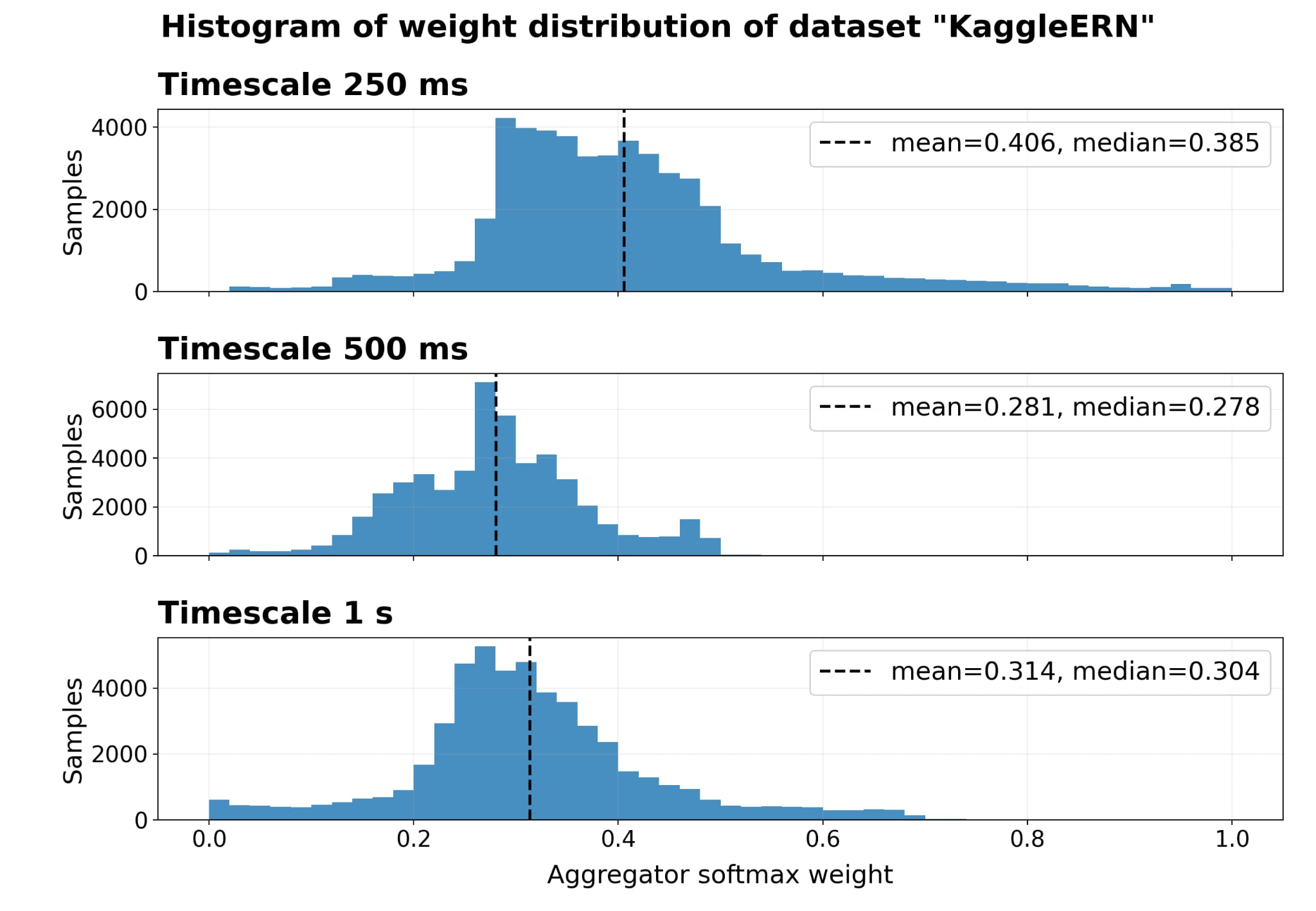}
    \caption{KaggleERN}
    \label{fig:kaggleern}
\end{subfigure}

\begin{subfigure}[b]{0.48\textwidth}
    \centering
    \includegraphics[width=\textwidth]{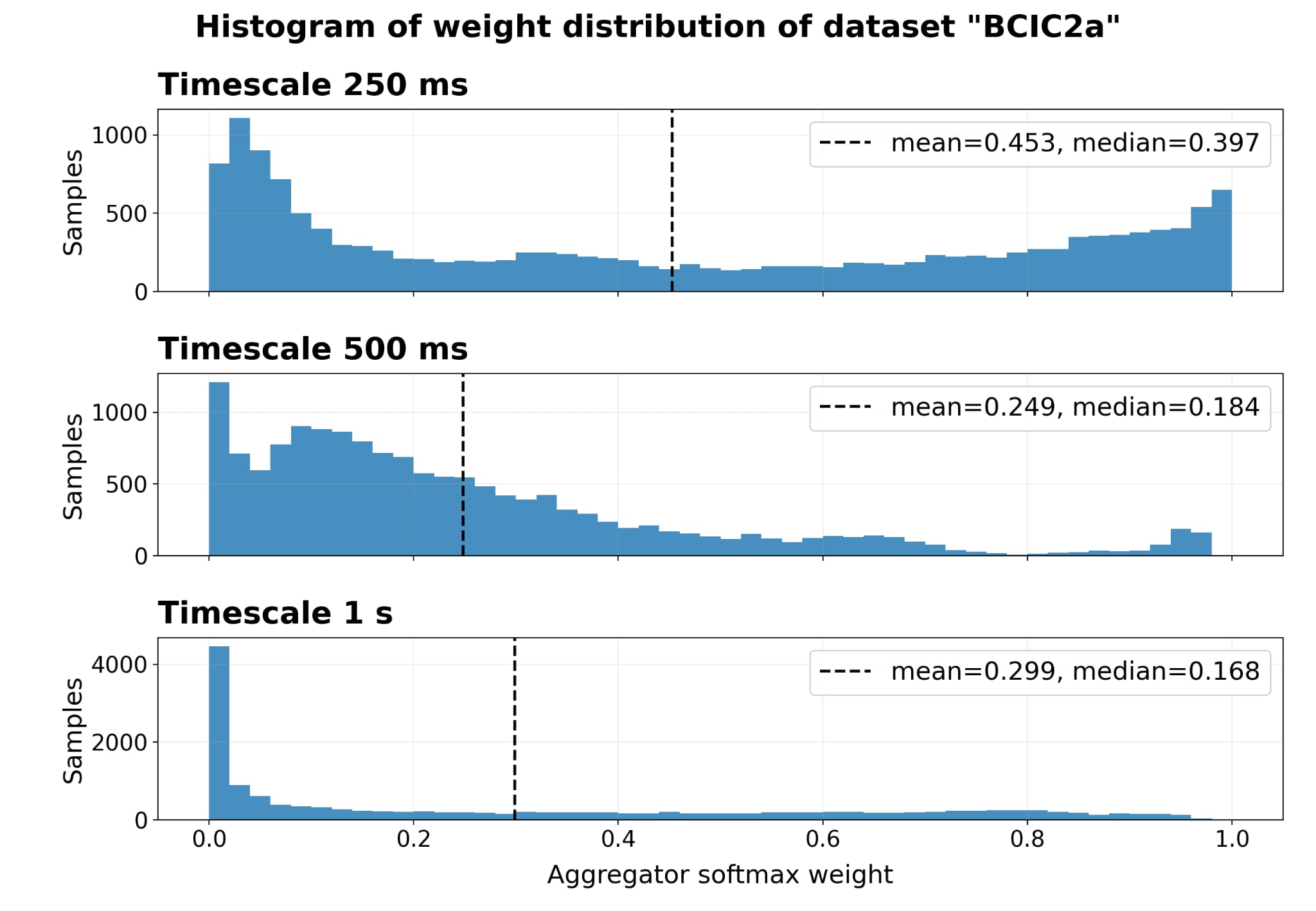}
    \caption{BCIC2a}
    \label{fig:bcic2a}
\end{subfigure}

\caption{Histogram of weight distribution of different datasets.}
\label{fig:histograms}
\end{figure}

In this section, we will discuss how the aggregator vary at the individual sample level within a single task. These results confirm that the aggregator dynamically adapts to individual signal characteristics rather than relying on a fixed per-task bias. The significant variance observed within a single fold proves that the routing weights cannot be predicted by dataset identity alone. If the aggregator were simply a task-level lookup, intra-task variance would be near zero. Instead, this dynamic adjustment to individual inputs confirms that MAGE performs routing based on signal content, rather than learning a fixed per-task bias (Figure~\ref{fig:histograms}).

\subsection{Ablation on adding noise during pretraining}
In this ablation, we study the effectiveness of adding noise during pretraining. Without noise, the model can collapse to using only one timescale, leaving the other unexplored. This typically occurs because an expert selected early on (due to random initialization) minimizes the loss faster than others. As shown in Table~\ref{tab:noisy_pretrain}, incorporating noise during the pretraining phase prevents this collapse and leads to better results. Specifically, this noise injection improves performance in three of the four evaluated datasets: FACED, Sleep-EDFx, and SHU-MI.

\begin{table}[htbp]
    \centering
    \caption{Effectiveness of Noise Injection on Preventing Granularity Collapse. MAGE pretrained with noise achieves a better performance. \textbf{Bold} indicates the best result. The result is reported on 1 seed.}
    \label{tab:noisy_pretrain}
    \begin{tabular}{llcc}
        \toprule
        Dataset & Metrics & MAGE & MAGE wo noise \\
        \midrule
        \multirow{3}{*}{FACED} & Balanced Acc & \textbf{51.30$_{\pm 3.11}$} & 50.35$_{\pm 2.62}$  \\
                               & Cohen's Kappa  & \textbf{44.89$_{\pm 3.43}$} & 43.80$_{\pm 3.02}$ \\
                               & Weighted F1    & \textbf{50.97$_{\pm 2.96}$} & 49.84$_{\pm 2.80}$ \\
        \midrule
        \multirow{3}{*}{PhysioNet-MI} & Balanced Acc & 61.15$_{\pm 0.86}$ & \textbf{61.57$_{\pm 1.15}$} \\
                               & Cohen's Kappa  & 48.23$_{\pm 1.15}$ & \textbf{48.78$_{\pm 1.51}$} \\
                               & Weighted F1   & 61.16$_{\pm 0.90}$ & \textbf{61.66$_{\pm 1.14}$}  \\
        \midrule
        \multirow{3}{*}{Sleep-EDFx} & Balanced Acc & \textbf{70.24$_{\pm 4.63}$} & 68.79$_{\pm 4.24}$ \\
                               & Cohen's Kappa  & \textbf{68.91$_{\pm 6.47}$} & 66.01$_{\pm 7.53}$ \\
                               & Weighted F1    & \textbf{77.31$_{\pm 5.00}$} & 75.59$_{\pm 5.72}$ \\
        \midrule
        \multirow{3}{*}{SHU-MI} & Balanced Acc & \textbf{62.43$_{\pm 3.18}$} & 62.24$_{\pm 4.24}$ \\
                               & AUC-PR  & \textbf{68.73$_{\pm 5.03}$} & 68.54$_{\pm 6.17}$ \\
                               & AUROC   & 68.14$_{\pm 5.06}$ & \textbf{68.45$_{\pm 5.53}$} \\
        \bottomrule
    \end{tabular}
\end{table}

\subsection{Ablation on different timescale levels}
\begin{figure*}
    \centering
    \includegraphics[width=0.99\linewidth]{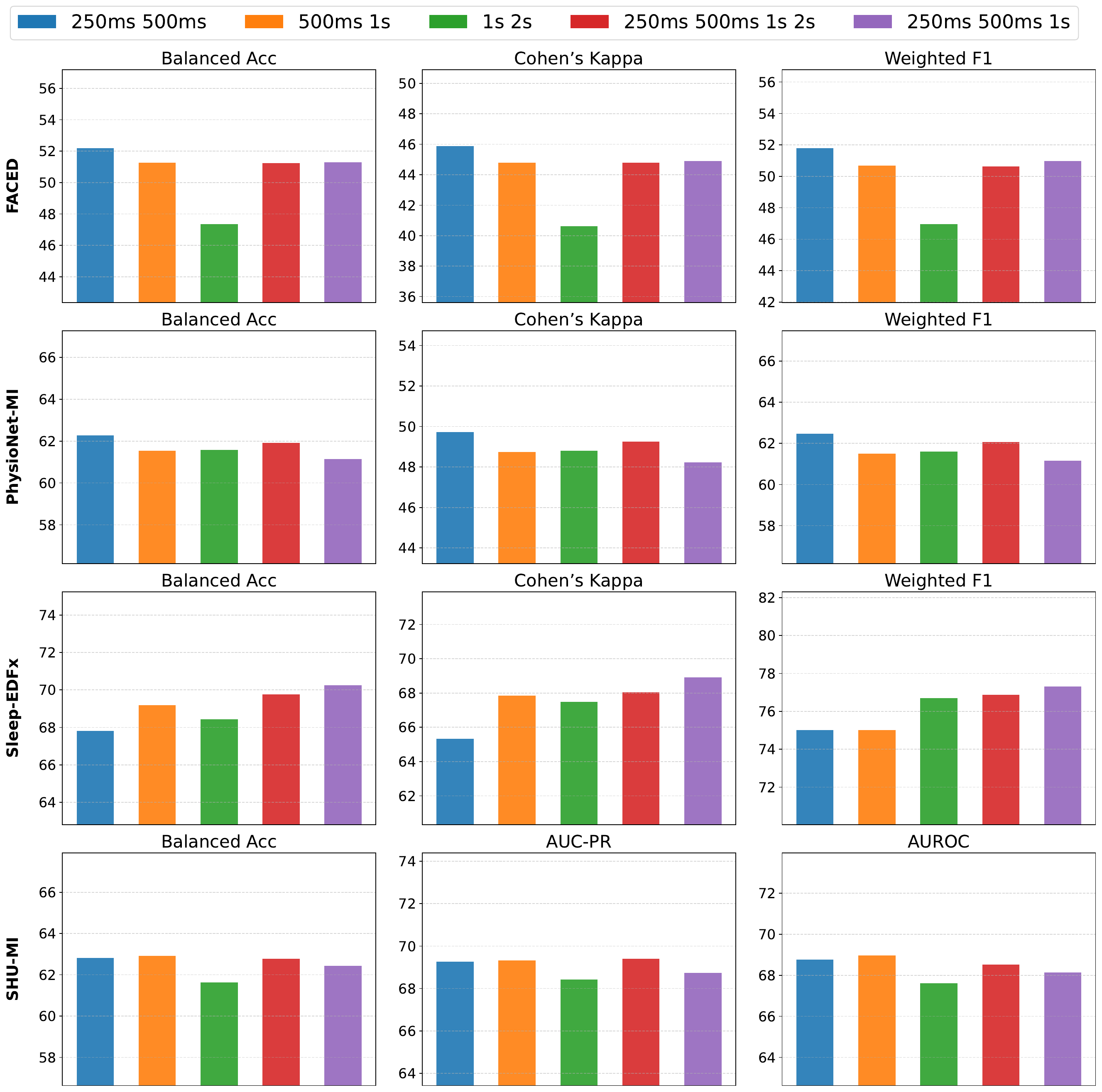}
    \caption{Impact of multi-timescale patch configurations on representation performance across different datasets. Modifying the patch size configurations results in varying performance levels across multiple datasets. The result is reported on 1 seed.}
    \label{fig:different_scale}
\end{figure*}

To investigate the impact of timescale on representation performance, we evaluated MAGE variants with patch configurations of [250ms, 500ms], [500ms, 1s], [1s, 2s], and [250ms, 500ms, 1s, 2s]. As illustrated in Fig.~\ref{fig:different_scale}, increasing the diversity of timescales results in a more generalizable representation, ensuring robust performance across all metrics and datasets. Notably, different patch sizes possess distinct inductive biases: while the [250ms, 500ms] configuration excels on FACED due to its reliance on short temporal windows, its performance on Sleep-EDFx is significantly degraded. Conversely, the coarser [1s, 2s] configuration achieves better performance on Sleep-EDFx than the [250ms, 500ms] configuration, but fails to generalize effectively to other datasets. In contrast, when employing configurations with three or four timescales, the performance remains consistently strong across all tasks, despite minor fluctuations on individual datasets. This demonstrates that a [250ms, 500ms, 1s] configuration is already sufficiently robust to achieve high generalizability.

\subsection{Ablation on Aggregator usage during pretraining}

This ablation investigates whether the Aggregator should be optimized during pretraining or introduced only during downstream adaptation. In the standard MAGE setting, the Aggregator is trained together with the multi-scale encoder during pretraining, allowing the model to learn task-agnostic scale fusion from the pretraining data. In contrast, the ablated variant performs uniform pretraining across all scales and initializes the Aggregator only during downstream fine-tuning.

As shown in Table~\ref{tab:uniform_pretrain_aggre_downstream}, using the Aggregator during pretraining leads to clear improvements on FACED and Sleep-EDFx. These results suggest that optimizing the Aggregator during pretraining helps MAGE learn more transferable multi-scale representations, especially for downstream tasks involving heterogeneous temporal and frequency patterns, such as emotion recognition and sleep staging.

\begin{table}
    \centering
    \caption{MAGE variant where the Aggregator is used only during downstream adaptation, while pretraining is performed with uniform training. The result is reported on 1 seed.}
    \label{tab:uniform_pretrain_aggre_downstream}
    \resizebox{0.7\linewidth}{!}{%
        \begin{tabular}{llcc}
            \toprule
            Dataset & Metrics & MAGE & MAGE-uniform pretrain \\
            \midrule
            \multirow{3}{*}{PhysioNet-MI} & Balanced Acc & 61.15$_{\pm 0.86}$ & 61.15$_{\pm 0.27}$ \\
                                          & Cohen's Kappa & 48.23$_{\pm 1.15}$ & 48.23$_{\pm 0.37}$ \\
                                          & Weighted F1   & 61.16$_{\pm 0.90}$ & 61.31$_{\pm 0.25}$ \\
            \midrule
            \multirow{3}{*}{FACED}      & Balanced Acc & 51.30$_{\pm 3.11}$ & 46.92$_{\pm 2.13}$ \\
                                        & Cohen's Kappa & 44.89$_{\pm 3.43}$ & 40.19$_{\pm 2.39}$ \\
                                        & Weighted F1   & 50.97$_{\pm 2.96}$ & 46.88$_{\pm 2.11}$ \\
            \midrule
            \multirow{3}{*}{Sleep-EDFx} & Balanced Acc & 70.24$_{\pm 4.63}$ & 63.07$_{\pm 4.84}$ \\
                                        & Cohen's Kappa & 68.91$_{\pm 6.47}$ & 63.49$_{\pm 7.62}$ \\
                                        & Weighted F1   & 77.31$_{\pm 5.00}$ & 72.61$_{\pm 5.70}$ \\
            \midrule
            \multirow{3}{*}{SHU-MI}     & Balanced Acc & 62.43$_{\pm 3.18}$ & 63.49$_{\pm 4.19}$ \\
                                        & AUC-PR       & 68.73$_{\pm 5.03}$ & 69.68$_{\pm 5.84}$ \\
                                        & AUROC        & 68.14$_{\pm 5.06}$ & 69.41$_{\pm 5.63}$ \\
            \bottomrule
        \end{tabular}%
    }
\end{table}

\subsection{Ablation on changing the ratio of alignment loss and reconstruction loss}
\begin{figure}
    \centering
    \includegraphics[width=0.99\linewidth]{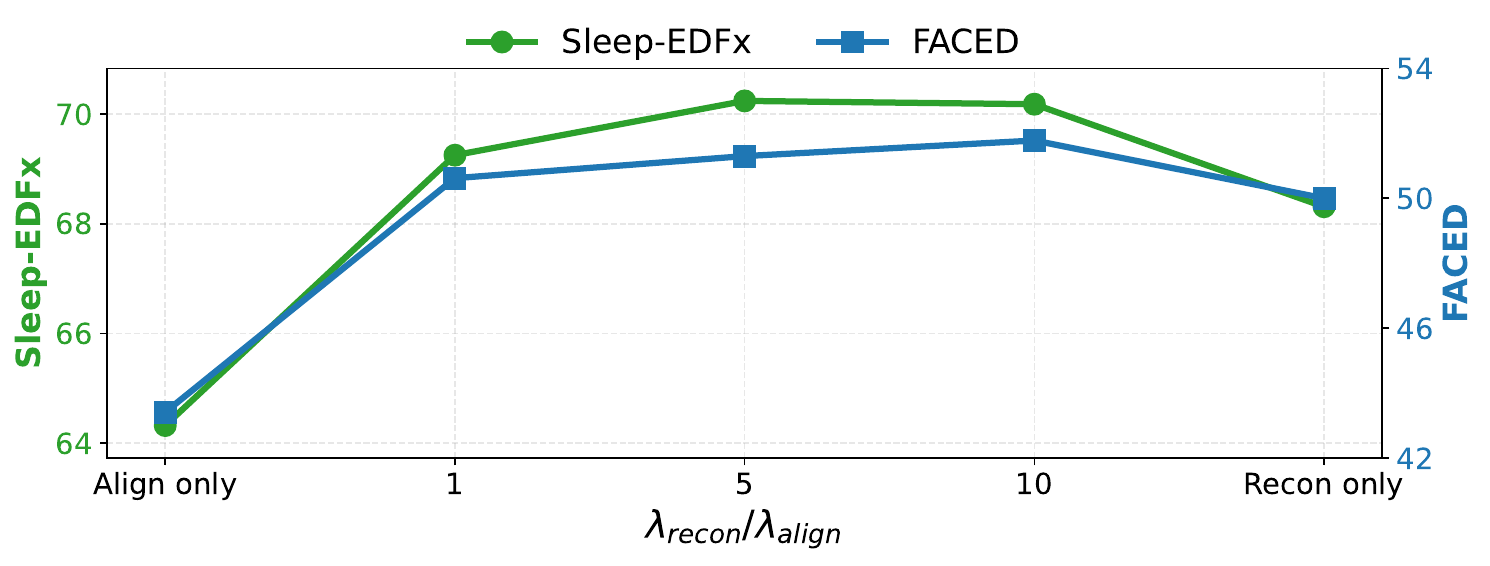}
    \caption{Balance accuracy of MAGE on two datasets when changing the ratio of $\lambda_\text{recon} / \lambda_\text{align}$. Using both types of loss achieves better performance than a single loss. The result is reported on 1 seed.}
    \label{fig:different_ratio}
\end{figure}

In this section, we investigate the individual contributions of the two objective functions used during the pre-training phase: alignment loss and reconstruction loss. Specifically, we conduct ablation studies by pre-training the model using only the alignment loss or only the reconstruction loss. Furthermore, we explore the effect of varying the corresponding hyperparameters, $\lambda_\text{recon}$ and $\lambda_\text{align}$, on the overall pre-training performance. The balanced accuracy for different ratios of these values on Sleep-EDFx and FACED is illustrated in Fig.~\ref{fig:different_ratio}. The results demonstrate that combining both loss functions consistently yields superior performance compared to relying on a single loss type. Crucially, optimizing with only the alignment loss leads to poor results, dropping to approximately $64\%$ balanced accuracy on Sleep-EDFx and $42\%$ on FACED. Furthermore, when both losses are utilized simultaneously, the performance differences across the various weight ratios remain relatively marginal.

\subsection{Ablation on changing the mask ratio}
\begin{table*}
\caption{Balanced accuracy on FACED of different pretraining masking ratio. The overall differences are marginal across different masking ratios. The result is reported on 1 seed. }
\label{tab:mask_ratio}
\centering

\resizebox{0.95\textwidth}{!}{%
    \begin{tabular}{llcccc}
    \toprule
    Dataset & Metrics & Mask $50\%$ & Mask $60\%$ & Mask $70\%$ & Mask $80\%$\\
    \midrule
    \multirow{3}{*}{FACED} & Balanced Acc & 50.83$_{\pm 3.84}$ & 51.22$_{\pm 3.44}$ & 51.30$_{\pm 3.11}$ & 51.30$_{\pm 2.21}$ \\
                                & Cohen's Kappa  & 44.31$_{\pm 4.24}$ & 44.83$_{\pm 3.77}$ & 44.89$_{\pm 3.43}$ & 44.80$_{\pm 2.49}$ \\
                                & Weighted F1    & 50.41$_{\pm 3.76}$ & 50.79$_{\pm 3.24}$ & 50.97$_{\pm 2.96}$ & 50.61$_{\pm 2.20}$ \\
    \bottomrule
    \end{tabular}%
}
\end{table*}

We run a small ablation study to verify the choice of the masking ratio. As shown in Table~\ref{tab:mask_ratio}, although downstream performance fluctuates slightly across different masking ratios, the overall differences are marginal, indicating that MAGE is highly robust to this hyperparameter. Consequently, we adopted a $70\%$ masking ratio as our default setting, consistent with some prior research.
\subsection{Ablation on using RoPE for channel information}

\begin{table*}
\caption{Comparison of using learnable channel embedding versus RoPE for channel positional embedding. The result is reported on 1 seed.}
\label{tab:rope_chan}
\centering

\resizebox{0.7\textwidth}{!}{%
    \begin{tabular}{llcc}
    \toprule
    Dataset & Metrics & Learnable Channel Embedding & RoPE\\
    \midrule
    \multirow{3}{*}{FACED} & Balanced Acc &  51.30$_{\pm 3.11}$ & 50.99$_{\pm 3.00}$ \\
                                & Cohen's Kappa  & 44.89$_{\pm 3.43}$ & 44.43$_{\pm 3.26}$\\
                                & Weighted F1    & 50.97$_{\pm 2.96}$ & 50.34$_{\pm 2.85}$ \\
    \bottomrule
    \end{tabular}%
}
\end{table*}

Unlike temporal data, EEG channels lack a natural linear order, making sequential bias from RoPE inappropriate. Instead, we employ trainable positional tokens (following BIOT and EEGPT) to enable the model to learn each electrode's unique neurophysiological role without assuming an artificial topology. Our small ablation study on FACED (Table~\ref{tab:rope_chan}) confirms this hypothesis: learnable tokens is consistently better than RoPE across all metrics, validating that specific channel identities are more informative than arbitrary sequences.

\subsection{Ablation on increasing the volume of pretraining hours}
\label{subsec: pretraining_hour}

\begin{table}
    \centering
    \caption{MAGE when pretrained on different training hours of EEG signal. The result is reported on 1 seed.}
    \label{tab:2500h}
    \resizebox{0.7\linewidth}{!}{%
        \begin{tabular}{llcc}
            \toprule
            Dataset & Metrics & MAGE-500h & MAGE-2500h \\
            \midrule
            \multirow{3}{*}{PhysioNet-MI} & Balanced Acc & 61.15$_{\pm 0.86}$ & \textbf{62.57$_{\pm 0.73}$} \\
                                          & Cohen's Kappa & 48.23$_{\pm 1.15}$ & \textbf{50.12$_{\pm 0.98}$} \\
                                          & Weighted F1   & 61.16$_{\pm 0.90}$ & \textbf{62.62$_{\pm 0.75}$} \\
            \midrule
            \multirow{3}{*}{FACED}      & Balanced Acc & 51.30$_{\pm 3.11}$ &  \textbf{52.02$_{\pm 2.22}$} \\
                                        & Cohen's Kappa & 44.89$_{\pm 3.43}$ & \textbf{45.68$_{\pm 2.40}$} \\
                                        & Weighted F1   & 50.97$_{\pm 2.96}$ & \textbf{51.62$_{\pm 1.94}$} \\
            \midrule
            \multirow{3}{*}{Sleep-EDFx} & Balanced Acc & \textbf{70.24$_{\pm 4.63}$} & 68.03$_{\pm 4.20}$ \\
                                        & Cohen's Kappa & \textbf{68.91$_{\pm 6.47}$} & 64.78$_{\pm 7.56}$ \\
                                        & Weighted F1   & \textbf{77.31$_{\pm 5.00}$} & 75.13$_{\pm 5.48}$ \\
            \midrule
            \multirow{3}{*}{SHU-MI}     & Balanced Acc & 62.43$_{\pm 3.18}$ & \textbf{63.56$_{\pm 4.11}$} \\
                                        & AUC-PR       & 68.73$_{\pm 5.03}$ & \textbf{68.99$_{\pm 6.68}$} \\
                                        & AUROC        & 68.14$_{\pm 5.06}$ & \textbf{69.10$_{\pm 5.80}$} \\
            \bottomrule
        \end{tabular}%
    }
\end{table}

In this section, we evaluate the impact of scaling up MAGE's pre-training data volume. We increased the dataset size fivefold to 2,500 hours, but this expansion yielded only minor improvements on certain tasks as shown in Table.~\ref{tab:2500h}. We believe that scaling the dataset further does not justify the massive increase in computational cost. This suggests that raw data volume is not the sole determinant of success; rather, the model's structural inductive bias is equally, if not more, critical. We, therefore, decide to pretrain on our subset of 500 hours for better trade-off between efficiency and performance.

\newpage
\section*{NeurIPS Paper Checklist}

\begin{enumerate}

\item {\bf Claims}
    \item[] Question: Do the main claims made in the abstract and introduction accurately reflect the paper's contributions and scope?
    \item[] Answer: \answerYes{} 
    \item[] Justification: The primary contribution of this paper is the adoption of a multi-scale architecture for EEG signal analysis, together with an aggregator that adaptively weights timescale importance for various downstream datasets. This contribution is accurately reflected in the abstract and introduction.

\item {\bf Limitations}
    \item[] Question: Does the paper discuss the limitations of the work performed by the authors?
    \item[] Answer: \answerYes{}.
    \item[] Justification: The Limitation is discussed in Section 4 of the paper.

\item {\bf Theory assumptions and proofs}
    \item[] Question: For each theoretical result, does the paper provide the full set of assumptions and a complete (and correct) proof?
    \item[] Answer: \answerNA{} 
    \item[] Justification: The paper does not provide any new theory assumptions or proof, it focuses on the architecture to enhance the generalizability of EEG foundation models.

    \item {\bf Experimental result reproducibility}
    \item[] Question: Does the paper fully disclose all the information needed to reproduce the main experimental results of the paper to the extent that it affects the main claims and/or conclusions of the paper (regardless of whether the code and data are provided or not)?
    \item[] Answer: \answerYes{} 
    \item[] Justification: All the experimental results can be reproduced by the information we provide in Section 4 and the Appendix. We additionally release the code if the paper is accepted. 

\item {\bf Open access to data and code}
    \item[] Question: Does the paper provide open access to the data and code, with sufficient instructions to faithfully reproduce the main experimental results, as described in supplemental material?
    \item[] Answer: \answerYes{} 
    \item[] Justification: We provide open access to our code, along with comprehensive instructions to reproduce our results. However, because our evaluation relies on open-source medical datasets containing sensitive health information, we are unable to distribute the raw data directly. Researchers wishing to use these datasets must independently apply for access through the original dataset providers.

\item {\bf Experimental setting/details}
    \item[] Question: Does the paper specify all the training and test details (e.g., data splits, hyperparameters, how they were chosen, type of optimizer) necessary to understand the results?
    \item[] Answer: \answerYes{} 
    \item[] Justification: We specify all training and test details in the Appendix, together with the ratio for splitting and the number of folds for each downstream dataset.

\item {\bf Experiment statistical significance}
    \item[] Question: Does the paper report error bars suitably and correctly defined or other appropriate information about the statistical significance of the experiments?
    \item[] Answer: \answerYes{} 
    \item[] Justification: We include error bars for our reported experiments. Because EEG signals inherently exhibit high inter-subject variability, our resulting error margins are correspondingly wide.

\item {\bf Experiments compute resources}
    \item[] Question: For each experiment, does the paper provide sufficient information on the computer resources (type of compute workers, memory, time of execution) needed to reproduce the experiments?
    \item[] Answer: \answerYes{} 
    \item[] Justification: We provide the information about the number of workers, as well as the GPU types we use during the experiments.
    
\item {\bf Code of ethics}
    \item[] Question: Does the research conducted in the paper conform, in every respect, with the NeurIPS Code of Ethics \url{https://neurips.cc/public/EthicsGuidelines}?
    \item[] Answer: \answerYes{} 
    \item[] Justification: The research conducted conforms to the NeurIPS Code of Ethics. We use only established, open-source EEG datasets and strictly adhere to their respective licensing and access agreements by not redistributing sensitive medical data. Furthermore, we ensure transparency and reproducibility by providing open access to our code and openly discussing the limitations of our study. 

\item {\bf Broader impacts}
    \item[] Question: Does the paper discuss both potential positive societal impacts and negative societal impacts of the work performed?
    \item[] Answer: \answerYes{} 
    \item[] Justification: Although this work represents an early, foundational step toward generalizable EEG modeling, we explicitly discuss its potential societal impacts in the future. We highlight the positive impacts of advancing clinical healthcare and BCI applications, as well as potential negative impacts, such as privacy concerns and the risks associated with the deployment of these models in real-world settings.
    
\item {\bf Safeguards}
    \item[] Question: Does the paper describe safeguards that have been put in place for responsible release of data or models that have a high risk for misuse (e.g., pre-trained language models, image generators, or scraped datasets)?
    \item[] Answer: \answerNA 
    \item[] Justification: Our model is a lightweight (0.6M parameter) feature extractor for EEG signals, rather than a generative model. Therefore, the paper does not describe safeguards.

\item {\bf Licenses for existing assets}
    \item[] Question: Are the creators or original owners of assets (e.g., code, data, models), used in the paper, properly credited and are the license and terms of use explicitly mentioned and properly respected?
    \item[] Answer: \answerYes{} 
    \item[] Justification: We cite the original creators of all baseline models and open-source EEG datasets evaluated in our work. Furthermore, we strictly respect the licenses and terms of use of these medical datasets.

\item {\bf New assets}
    \item[] Question: Are new assets introduced in the paper well documented and is the documentation provided alongside the assets?
    \item[] Answer: \answerYes{} 
    \item[] Justification: We will make the codebase open access when the paper is accepted. We additionally provide a README, which includes step-by-step instructions on how to set up the environment, run the models, and fully reproduce the experimental results.

\item {\bf Crowdsourcing and research with human subjects}
    \item[] Question: For crowdsourcing experiments and research with human subjects, does the paper include the full text of instructions given to participants and screenshots, if applicable, as well as details about compensation (if any)? 
    \item[] Answer: \answerNA{} 
    \item[] Justification: This paper does not involve any new crowdsourcing experiments or primary data collection with human subjects.

\item {\bf Institutional review board (IRB) approvals or equivalent for research with human subjects}
    \item[] Question: Does the paper describe potential risks incurred by study participants, whether such risks were disclosed to the subjects, and whether Institutional Review Board (IRB) approvals (or an equivalent approval/review based on the requirements of your country or institution) were obtained?
    \item[] Answer: \answerNA{} 
    \item[] Justification: We do not conduct experiments on new study participants.

\item {\bf Declaration of LLM usage}
    \item[] Question: Does the paper describe the usage of LLMs if it is an important, original, or non-standard component of the core methods in this research? Note that if the LLM is used only for writing, editing, or formatting purposes and does \emph{not} impact the core methodology, scientific rigor, or originality of the research, declaration is not required.
    \item[] Answer: \answerNA{} 
    \item[] Justification:  LLM is used only for writing, editing, or formatting purposes and does not impact the core methodology, scientific rigor, or originality of the research.

\end{enumerate}

\end{document}